\documentclass[10pt,twocolumn,letterpaper]{article}

\usepackage{cvpr}              

\usepackage{graphicx}
\usepackage{amsmath}
\usepackage{amssymb}
\usepackage{booktabs}
\usepackage{multirow}
\usepackage{float}
\usepackage{tabularx}
\makeatletter
\@namedef{ver@everyshi.sty}{}
\makeatother
\usepackage{tikz}
\usepackage{comment}
\usepackage{color}
\usepackage{amsfonts}
\usepackage{caption}
\usepackage{mathtools}
\usepackage{tablefootnote}
\usepackage{dsfont}
\usepackage{ragged2e}
\usepackage{array}

\usepackage[pagebackref,breaklinks,colorlinks]{hyperref}

\newcommand{\OURS}{RoITr}

\usepackage[capitalize]{cleveref}
\crefname{section}{Sec.}{Secs.}
\Crefname{section}{Section}{Sections}
\Crefname{table}{Table}{Tables}
\crefname{table}{Tab.}{Tabs.}

\def\cvprPaperID{1048} 
\def\confName{CVPR}
\def\confYear{2023}

\begin{document}

\title{Rotation-Invariant Transformer for Point Cloud Matching}

\author{Hao Yu$^1$ \; Zheng Qin$^2$ \;  Ji Hou$^{1}$\; Mahdi Saleh$^{1, 3}$ \; Dongsheng Li$^2$ \; Benjamin Busam$^{1, 3}$ \; Slobodan Ilic$^{1, 4}$ \\
$^{1}$TU Munich \quad $^{2}$NUDT \quad $^{3}$3Dwe.ai \quad$^{4}$Siemens AG, Munich
}

\maketitle
\begin{abstract}
The intrinsic rotation invariance lies at the core of matching point clouds with handcrafted descriptors. However, it is widely despised by recent deep matchers that obtain the rotation invariance extrinsically via data augmentation. As the finite number of augmented rotations can never span the continuous $SO(3)$ space, these methods usually show instability when facing rotations that are rarely seen. To this end, we introduce \OURS{}, a \textbf{Ro}tation-\textbf{I}nvariant \textbf{Tr}ansformer to cope with the pose variations in the point cloud matching task. We contribute both on the local and global levels.
Starting from the local level, we introduce an attention mechanism embedded with Point Pair Feature~(PPF)-based coordinates to describe the pose-invariant geometry, upon which a novel attention-based encoder-decoder architecture is constructed. We further propose a global transformer with rotation-invariant cross-frame spatial awareness learned by the self-attention mechanism, which significantly improves the feature distinctiveness and makes the model robust with respect to the low overlap. Experiments are conducted on both the rigid and non-rigid public benchmarks, where \OURS{} outperforms all the state-of-the-art models by a considerable margin in the low-overlapping scenarios. Especially when the rotations are enlarged on the challenging 3DLoMatch benchmark, \OURS{} surpasses the existing methods by at least 13 and 5 percentage points in terms of \textit{Inlier Ratio} and \textit{Registration Recall}, respectively. Code is publicly available~\footnote{\href{https://github.com/haoyu94/RoITr}{https://github.com/haoyu94/RoITr}}.
\end{abstract}
\vspace{-0.5cm}
\section{Introduction}
\label{sec:intro}
The correspondence estimation between a pair of partially-overlapping point clouds is a long-standing task that lies at the core of many computer vision applications, such as tracking~\cite{huang2016volumetric,huang2017tracking}, reconstruction~\cite{newcombe2011kinectfusion,newcombe2015dynamicfusion,tang2021learning}, pose estimation~\cite{huang2021predator,yu2021cofinet,qin2022geometric} and 3D representation learning~\cite{xie2020pointcontrast, hou2021exploring,hou2021pri3d}, etc. In a typical solution, geometry is first encoded into descriptors, and correspondences are then established between two frames by matching the most similar descriptors. As the two frames are observed from different views, depicting the same geometry under different transformations identically, \ie., the pose-invariance, becomes the key to success in the point cloud matching task. 
\begin{figure}
  \includegraphics[width=0.48\textwidth]{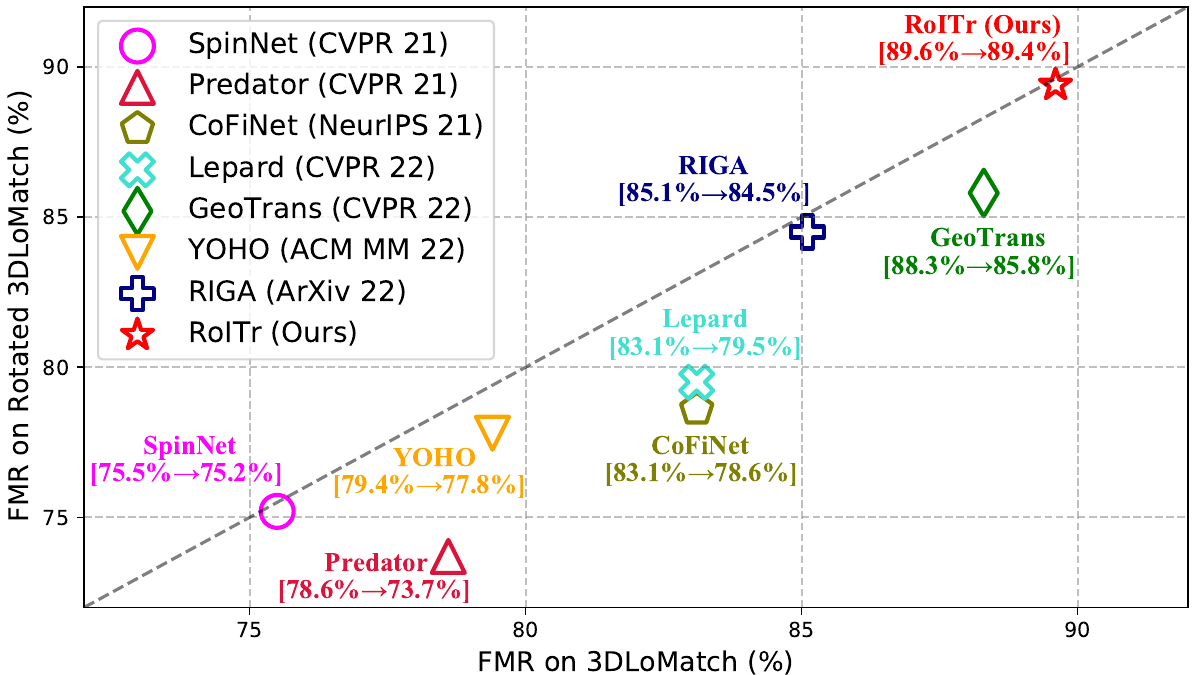}
  \vspace{-0.6cm}
  \caption{\textit{Feature Matching Recall}~(FMR) on 3DLoMatch~\cite{huang2021predator} and Rotated 3DLoMatch. Distance to the diagonal represents the robustness against rotations. Among all the state-of-the-art approaches, \OURS{} not only ranks first on both benchmarks but also shows the best robustness against the enlarged rotations.}
  \label{fig:teaser}
  \vspace{-0.6cm}
\end{figure}

Since the side effects caused by a global translation can always be easily eliminated, \eg., by aligning the barycenter with the origin, the attention naturally shifts to coping with the rotations. 
In the past, handcrafted local descriptors~\cite{rusu2008aligning,rusu2009fast,drost2010model,tombari2010unique} were designed to be rotation-invariant so that the same geometry observed from different views can be correctly matched. With the emergence of deep neural models for 3D point analysis, \eg., multilayer perceptrons~(MLPs)-based like PointNet~\cite{qi2017pointnet,qi2017pointnet++}, convolutions-based like KPConv~\cite{thomas2019kpconv,choy20194d}, and the attention-based like PointTransformer~\cite{zhao2021point,saleh2022cloudattention}, recent approaches~\cite{zeng20173dmatch,deng2018ppfnet,deng2018ppf,gojcic2019perfect,choy2019fully,saleh2020graphite,ao2021spinnet,huang2021predator,yu2021cofinet,qin2022geometric,li2022lepard,yew2022regtr,yu2022riga} propose to learn descriptors from raw points as an alternative to handcrafted features that are less robust to occlusion and noise. The majority of deep point matchers~\cite{zeng20173dmatch,deng2018ppfnet,choy2019fully,huang2021predator,yu2021cofinet,saleh2022bending,qin2022geometric,li2022lepard,yew2022regtr,zhang2022pcr} is sensitive to rotations. Consequently, their invariance to rotations  must be obtained extrinsically via augmented training to ensure that the same geometry under different poses can be depicted similarly.  However, as the training cases can never span the continuous $SO(3)$ space, they always suffer from instability when facing rotations that are rarely seen during training. This can be observed by a significant performance drop under enlarged rotations at inference time.~(See Fig.~\ref{fig:teaser}.) 

There are other works~\cite{deng2018ppf,gojcic2019perfect,saleh2020graphite,ao2021spinnet,wang2022you} that only leverage deep neural networks to encode the pure geometry with the intrinsically-designed rotation invariance.  However, the intrinsic rotation invariance comes at the cost of losing global context. For example, a human's left and right halves are almost identically described, which naturally degrades the distinctiveness of features. Most recently, RIGA~\cite{yu2022riga} is proposed to enhance the distinctiveness of the rotation-invariant descriptors by incorporating a global context, \eg., the left and right halves of a human become distinguishable by knowing there is a chair on the left while a table on the right. 
However, it lacks a highly-representative geometry encoder since it relies on PointNet~\cite{qi2017pointnet}, which accounts for an ineffective local geometry description.
Moreover, as depicting the cross-frame spatial relationships is non-trivial, previous works~\cite{huang2021predator,yu2021cofinet,saleh2022bending,qin2022geometric} merely leverage the contextual features in the cross-frame context aggregation, which neglects the positional information. Although RIGA proposes to learn a rotation-invariant position representation by leveraging an additional PointNet, this simple design is hard to model the complex cross-frame positional relationships and leads to less distinctive descriptors.

In this paper, we present \textbf{Ro}tation-\textbf{I}nvariant \textbf{Tr}ansformer (\OURS{}) to tackle the problem of point cloud matching under arbitrary pose variations. 
By using Point Pair Features~(PPFs) as the local coordinates, we propose an attention mechanism to learn the pure geometry regardless of the varying poses. Upon it, attention-based layers are further proposed to compose the encoder-decoder architecture for highly-discriminative and rotation-invariant geometry encoding. We demonstrate its superiority over PointTransformer~\cite{zhao2021point}, a state-of-the-art attention-based backbone network, in terms of both efficiency and efficacy in Fig.~\ref{fig:runtime} and Tab.~\ref{tab:ablation}~(a), respectively. On the global level, the cross-frame position awareness is introduced in a rotation-invariant fashion to facilitate feature distinctiveness. We illustrate its significance over the state-of-the-art design~\cite{qin2022geometric} in Tab.~\ref{tab:ablation}~(d). Our main contributions are summarized as:
\begin{itemize}

\item An attention mechanism designed to disentangle the geometry and poses, which enables the pose-agnostic geometry description.
\vspace{-0.2cm}
\item An attention-based encoder-decoder architecture that learns highly-representative local geometry in a rotation-invariant fashion.
\vspace{-0.2cm}
\item A global transformer with rotation-invariant cross-frame position awareness that significantly enhances the feature distinctiveness. 

\end{itemize}
\vspace{-0.3cm}
\section{Related Work}
\label{sec:related}

\noindent\textbf{Models with Extrinsic Rotation Invariance.} The mainstream of deep learning-based point cloud matching approaches is intrinsically rotation-sensitive. Pioneers \cite{zeng20173dmatch,deng2018ppfnet} learn to describe local patches from a rotation-variant input. FCGF~\cite{choy2019fully} leverages fully-convolutional networks to accelerate the geometry description. D3Feat~\cite{bai2020d3feat} jointly detects and describes sparse keypoints for matching. Predator~\cite{huang2021predator} incorporates the global context to enhance the local descriptors and predicts the overlap regions for keypoint sampling. CoFiNet~\cite{yu2021cofinet} extracts coarse-to-fine correspondences to alleviate the repeatability issue of keypoints. GeoTrans~\cite{qin2022geometric} considers the geometric information in fusing the intra-frame context globally. However, the awareness of spatial positions is missing in the cross-frame aggregation. Lepard~\cite{li2022lepard} extends the non-rigid shape matching~\cite{trappolini2021shape,saleh2022bending,tang2022neural} to point clouds~\cite{qin2023deep} and proposes a re-positioning module to alleviate the pose variations. REGTR~\cite{yew2022regtr} directly regresses the corresponding coordinates and registers point clouds in an end-to-end fashion. Nonetheless, all of these methods suffer from instability with additional rotations.

\noindent\textbf{Methods with Intrinsic Rotation Invariance.} A branch of handcrafted descriptors~\cite{chua1997point, tombari2010unique, guo2013rotational} aligns the input to a canonical representation according to an estimated local reference frame~(LRF), while the others~\cite{rusu2008aligning,rusu2009fast,drost2010model} mine the rotation-invariant components and encode them as the representation of the local geometry. Inspired by that, some deep learning-based methods~\cite{deng2018ppf,gojcic2019perfect,barroso2020hdd,saleh2020graphite,ao2021spinnet,yu2022riga,saleh2022bending} are designed to be intrinsically rotation-invariant to make the neural models focus on the pose-agnostic pure geometry. As a pioneer, PPF-FoldNet~\cite{deng2018ppf} consumes PPF-based patches and learns the descriptors using a FoldingNet~\cite{yang2018foldingnet}-based architecture without supervision. LRF-based works~\cite{gojcic2019perfect,saleh2020graphite,ao2021spinnet,saleh2022bending} achieve rotation invariance by aligning their input to the defined canonical representation. YOHO~\cite{wang2022you} adopts a group of rotations to learn a rotation-equivariant feature group and further obtain the invariance via group pooling. A common problem of the rotation-invariant methods is the less distinctive features. Although RIGA~\cite{yu2022riga} incorporates the global context into local descriptors to enhance the feature distinctiveness, its ineffective local geometry encoding and global position description learned by PointNet~\cite{qi2017pointnet} still constraint the representation ability of its descriptors.

 \begin{figure*}
  \includegraphics[width=\textwidth]{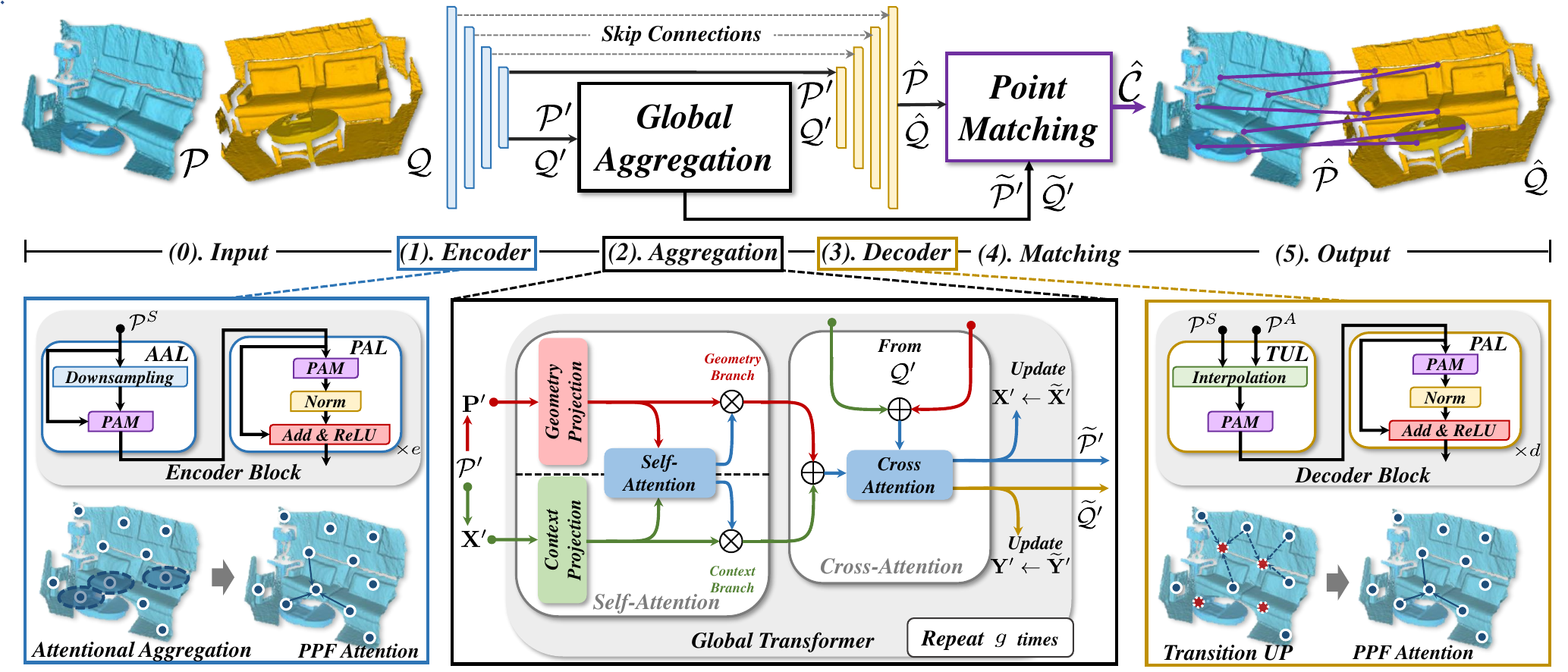}
  \vspace{-0.5cm}
  \caption{\textbf{An Overview of \OURS{}.} From left to right: \textit{\textbf{(0).}} \OURS{} takes as input a pair of triplets $\mathcal{P} = (\mathbf{P}, \mathbf{N}, \mathbf{X})$ and $\mathcal{Q} = (\mathbf{Q}, \mathbf{M}, \mathbf{Y})$, each with three dimensions referring to the point cloud, the estimated normals, and the initial features. \textit{\textbf{(1).}}[$\S{}$.~\ref{sec:local_geometry}] A stack of encoder blocks hierarchically downsamples the points to coarser superpoints and encodes the local geometry, yielding superpoint triplets $\mathcal{P}^\prime$ and $\mathcal{Q}^\prime$. Each encoder block consists of an Attentional Abstraction Layer~(AAL) for downsampling and abstraction, followed by $e\times$ PPF Attention Layers~(PALs) for local geometry encoding and context aggregation. Both of them are based on our proposed PPF Attention Mechanism~(PAM), which enables the pose-agnostic encoding of pure geometry.~(See Fig.~\ref{fig:differences} and Fig.~\ref{fig:local_attention}). \textit{\textbf{(2).}}[$\S{}$.~\ref{sec:global_context}] Global information is fused to enhance the superpoint features of $\mathcal{P}^\prime$ and $\mathcal{Q}^\prime$. The geometric cues are globally aggregated as a rotation-invariant position representation, which introduces spatial awareness in the consecutive cross-frame context aggregation. After a stack of $g\times$ global transformers, the globally-enhanced triplets $\widetilde{\mathcal{P}}^\prime$ and  $\widetilde{\mathcal{Q}}^\prime$ are produced. \textit{\textbf{(3).}}[$\S{}$.~\ref{sec:local_geometry}] Superpoint triplets $\mathcal{P}^\prime$ and $\mathcal{Q}^\prime$ are decoded to point triplets $\hat{\mathcal{P}}$ and $\hat{\mathcal{Q}}$ by a stack of decoder blocks. Each block consists of a Transition Up Layer~(TUL) for upsampling and context aggregation, followed by $d\times$ PALs.  \textit{\textbf{(4).}}[$\S{}$.~\ref{sec:matching}] By adopting the coarse-to-fine matching~\cite{yu2021cofinet}, $\widetilde{\mathcal{P}}^\prime$ and $\widetilde{\mathcal{Q}}^\prime$ are matched to generate superpoint correspondences, which are consecutively refined to point correspondences between $\hat{\mathcal{P}}$ and $\hat{\mathcal{Q}}$. \textit{\textbf{(5).}} $\hat{\mathcal{C}}$ is established between $\hat{\mathcal{P}}$ and $\hat{\mathcal{Q}}$.}
  \label{fig:pipeline}
  \vspace{-0.5cm}
\end{figure*}

\vspace{-0.2cm}
\section{Method}
\label{sec:method}
\noindent\textbf{Problem Statement.} We tackle the problem of matching a pair of partially-overlapping point clouds $\mathbf{P}\in \mathbb{R}^{n \times 3}$ and $\mathbf{Q}\in \mathbb{R}^{m \times 3}$, and extract a correspondence set $\hat{\mathcal{C}}=\{(\hat{\mathbf{p}}_{i}, \hat{\mathbf{q}}_{j})\big| \hat{\mathbf{p}}_i\in \hat{\mathbf{P}} \subseteq \mathbf{P}, \hat{\mathbf{q}}_j\in \hat{\mathbf{Q}} \subseteq \mathbf{Q}\}$ that minimizes:
\vspace{-0.1cm}
\begin{equation}
\frac{1}{|\hat{\mathcal{C}}|} \sum_{(\hat{\mathbf{p}}_{i}, \hat{\mathbf{q}}_{j})\in \hat{\mathcal{C}}} \lVert \mathcal{M}^*(\hat{\mathbf{p}}_{i}) - \hat{\mathbf{q}}_{j} \rVert_2,
\label{eq:definition}
\vspace{-0.2cm}
\end{equation}
where $\lVert\cdot\rVert_2$ denotes the Euclidean norm and $|\cdot|$ is the set cardinality. $\mathcal{M}^*(\cdot)$ stands for the ground-truth mapping function that maps $\hat{\mathbf{p}}_i$ to its corresponding position in $\hat{\mathbf{Q}}$. In rigid scenarios, it is defined by a transformation $\mathbf{T}^*\in SE(3)$. For the non-rigid cases it can be denoted as a per-point flow $\mathbf{f}^*_i\in\mathbb{R}^{3}$ known as the deformation field.

\noindent\textbf{Method Overview.} An overview of \OURS{} is shown in Fig.~\ref{fig:pipeline}. \OURS{} consists of an encoder-decoder architecture named Point Pair Feature Transformer~(PPFTrans) for local geometry encoding and a stack of $g\times$ global transformers for global context aggregation. Correspondence set $\hat{\mathcal{C}}$ is extracted by the coarse-to-fine matching~\cite{yu2021cofinet}.

\subsection{PPF Attention Mechanism}

\begin{figure}
  \includegraphics[width=0.48\textwidth]{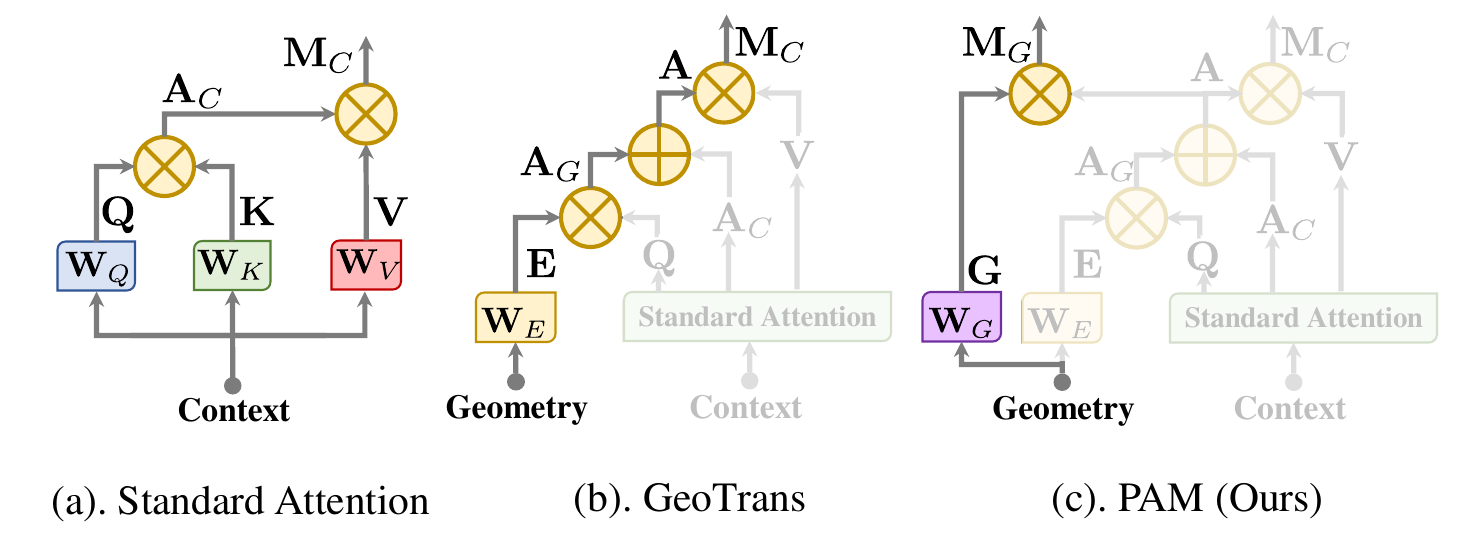}
  \caption{Illustration of different self-attention computation in the standard attention~\cite{vaswani2017attention}, GeoTrans~\cite{qin2022geometric}, and PAM.}
  \label{fig:differences}
  \vspace{-0.5cm}
\end{figure}

\noindent\textbf{Overview.} Fig.~\ref{fig:differences} compares three different self-attention mechanisms. The standard attention~\cite{vaswani2017attention} only leverages the input context to obtain the \textit{Query} $\mathbf{Q}$ and \textit{Key} $\mathbf{K}$ to compute the contextual attention $\mathbf{A}_C$, as well as the \textit{Value} $\mathbf{V}$ that encodes information for the contextual message $\mathbf{M}_C$. GeoTrans~\cite{qin2022geometric} proposes to learn the positional encoding $\mathbf{E}$ from the geometry and calculates a second attention $\mathbf{A}_G$ to reweigh $\mathbf{A}_C$. However, the cues contained in the raw geometry are totally neglected. To this end, we propose to learn the pose-agnostic geometric cues $\mathbf{G}$ and further generate the geometric message $\mathbf{M}_G$ in the PPF Attention Mechanism~(PAM). On the local level, $\mathbf{M}_G$ is combined with $\mathbf{M}_C$ for feature enhancement, while on the global level, it is used to learn the rotation-invariant position representation for the cross-frame context aggregation.
More specifically, we define PAM on an \textit{Anchor} triplet $\mathcal{P}^{A}=(\mathbf{P}^{A}, \mathbf{N}^{A}, \mathbf{X}^{A})$ and a \textit{Support} triplet $\mathcal{P}^{S}=(\mathbf{P}^{S}, \mathbf{N}^{S}, \mathbf{X}^{S})$, both with three dimensions referring to the point cloud, the estimated normals, and the associated features, respectively. PAM aggregates the learned context and geometric cues from $\mathcal{P}^{S}$ and flows the messages to $\mathcal{P}^{A}$.

\noindent\textbf{Pose-Agnostic Coordinate Representation.} The basis of PAM is the pose-agnostic local coordinate representation that we construct based on PPFs~\cite{drost2010model}. Let $\mathcal{P}^{A}_i := (\mathbf{p}^{A}_i\in \mathbf{P}^{A}, \mathbf{n}^{A}_i\in \mathbf{N}^{A}, \mathbf{x}^{A}_i\in \mathbf{X}^{A})\in \mathcal{P}^{A}$ denote the triplet constructed by picking the $i^{th}$ item on each dimension. For each $\mathbf{p}^{A}_i$, a subset of $\mathcal{P}^{S}$ is first retrieved according to the Euclidean distance w.r.t. $\mathbf{P}^S$, denoted as $\mathcal{P}^{S}_{\mathcal{N}(i)} := (\mathbf{P}^{S}_{\mathcal{N}(i)}, \mathbf{N}^{S}_{\mathcal{N}(i)},\mathbf{X}^{S}_{\mathcal{N}(i)})\subseteq \mathcal{P}^{S}$, with $\mathcal{N}(i)$ the indices of $k$-nearest neighbors. We then adopt PPFs~\cite{drost2010model} to construct a local coordinate system around each $\mathbf{p}^{A}_i$ to represent the pose-agnostic position of $\mathbf{P}^{S}_{\mathcal{N}(i)}$ w.r.t. it. The coordinate of point $\mathbf{p}_j^{S} \in \mathbf{P}^{S}_{\mathcal{N}(i)}$ is transferred to:
\begin{equation}
 \mathbf{e}^{S}_j = (\lVert\mathbf{d}\rVert_2, \angle(\mathbf{n}^{A}_i, \mathbf{d}), \angle(\mathbf{n}_j^{S}, \mathbf{d}), \angle(\mathbf{n}_j^{S}, \mathbf{n}^{A}_i)),
\label{eq:ppfs}
\end{equation}
with $\mathbf{d} = \mathbf{p}_j^{S} - \mathbf{p}_i^{A}$, and $\mathbf{n}_i^{A}$ and $\mathbf{n}_j^{S}$ the estimated normals of $\mathbf{p}_i^{A}$ and $\mathbf{p}_j^{S}$, respectively. $\angle(\mathbf{v}_1,\mathbf{v}_2)$ computes the angles between the two vectors~\cite{deng2018ppf,birdal2015point}. The transferred coordinates of $\mathbf{P}^{S}_{\mathcal{N}(i)}$ are denoted as $\mathbf{E}^{S}_{\mathcal{N}(i)}$.

\begin{figure}
  \includegraphics[width=0.48\textwidth]{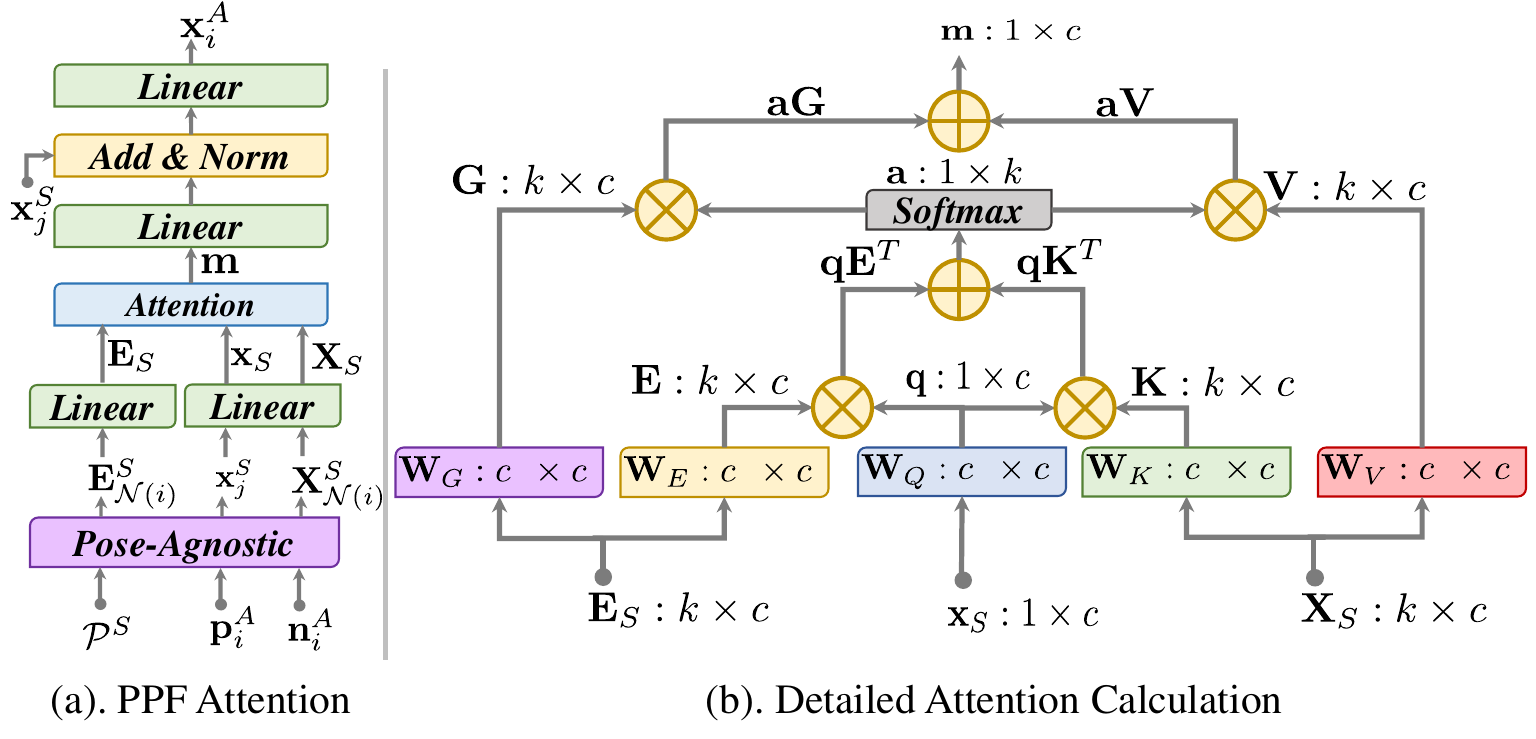}
  \caption{\textbf{Left}: The workflow of the PPF Attention Mechanism~(PAM). \textbf{Right}: Detailed calculation of the attention.}
  \label{fig:local_attention}
  \vspace{-0.5cm}
\end{figure}

\noindent\textbf{PPF Attention Mechanism.} 
PPF Attention Mechanism~(PAM) takes as input the \textit{Support} triplet $\mathcal{P}^{S}$ and the \textit{Anchor} point cloud $\mathbf{P}^{A}$ with estimated normals $\mathbf{N}^{A}$. PAM generates the \textit{Anchor} features $\mathbf{X}^{A}$ by aggregating the pose-agnostic local geometry and highly-representative learned context from $\mathcal{P}^{S}$, which is defined as:

\begin{equation}
\mathcal{P}^{A} = \delta(\mathbf{P}^{A}, \mathbf{N}^{A} \big| \mathcal{P}^{S}),
\label{eq:pam}
\end{equation}

\noindent with $\delta(\cdot)$ representing PAM. As shown in Fig.~\ref{fig:local_attention}~(a), for each $\mathbf{p}^A_i\in \mathbf{P}^A$ with normal $\mathbf{n}^A_i$, we find its nearest point $\mathbf{p}^S_j\in \mathbf{P}^S$ whose associated feature $\mathbf{x}^S_j$ is assigned to $\mathbf{p}^A_i$ as the initial description. Then, $k$-nearest neighbors from $\mathbf{P}^{S}$ are retrieved according to the Euclidean distance in 3D space, yielding $\mathbf{P}^{S}_{\mathcal{N}(i)}\subseteq\mathbf{P}^{S}$ and $\mathbf{X}^{S}_{\mathcal{N}(i)}\subseteq\mathbf{X}^{S}$. Following Eq.~\ref{eq:ppfs}, $\mathbf{P}^{S}_{\mathcal{N}(i)}$ is transferred to the pose-agnostic position representation $\mathbf{E}^{S}_{\mathcal{N}(i)}$, which is consecutively projected to the coordinate embedding $\mathbf{E}_S$ via a linear layer. ${\mathbf{x}}^{S}_j$ and $\mathbf{X}^{S}_{\mathcal{N}(i)}$ are projected to the contextual features $\mathbf{x}_S$ and $\mathbf{X}_S$ by a second shared linear layer, respectively. In Fig.~\ref{fig:local_attention}~(b), the attention mechanism uses five learnable matrices $\mathbf{W}_G$, $\mathbf{W}_E$, $\mathbf{W}_Q$, $\mathbf{W}_K$, and $\mathbf{W}_V$ to project the input. Specifically, $\mathbf{W}_G$ and $\mathbf{W}_E$ project the input coordinate representation to the geometric cues and positional encoding by:

\begin{equation}
    \mathbf{G} = \mathbf{E}_S\mathbf{W}_G \quad \text{and} \quad \mathbf{E} = \mathbf{E}_S\mathbf{W}_E,
\label{eq:project1}
\end{equation}

\noindent respectively. Similarly, $\mathbf{W}_Q$, $\mathbf{W}_K$, and $\mathbf{W}_V$ project the learned context to \textit{Query}, \textit{Key}, and \textit{Value} as:

\vspace{-0.5cm}
\begin{equation}
    \mathbf{q} = \mathbf{x}_S\mathbf{W}_Q, \; \mathbf{K} = \mathbf{X}_S\mathbf{W}_K, \; \text{and} \; \mathbf{V} = \mathbf{X}_S\mathbf{W}_V,
\label{eq:project2}
\vspace{-0.2cm}
\end{equation}
respectively. The attention $\mathbf{a}$ that measures the feature similarity, and the message $\mathbf{m}$ that encodes both the pose-agnostic geometry and the representative context read as:
\begin{equation}
    \mathbf{a} = \text{Softmax}(\frac{\mathbf{q}\mathbf{E}^{T} + \mathbf{q}\mathbf{K}^{T}}{\sqrt{c_0}}) \quad \text{and} \quad \mathbf{m} = \mathbf{a}\mathbf{G} + \mathbf{a}\mathbf{V},
\label{eq:local_attention}
\end{equation}
respectively. The message $\mathbf{m}$ is projected and aggregated to $\mathbf{x}^S_j$ via an element-wise addition followed by a normalization through LayerNorm~\cite{ba2016layer}. The final linear layer projects the obtained feature to $\mathbf{x}^{A}_i$, from which $\mathbf{X}^{A}$ is obtained to formulate the output $\mathcal{P}^A$ with the known $\mathbf{P}^A$ and $\mathbf{N}^A$.

\subsection{PPFTrans for Local Geometry Description}
\label{sec:local_geometry}
\noindent\textbf{Overview.} As illustrated in Fig.~\ref{fig:pipeline}, PPFTrans consumes triplets $\mathcal{P}$ and $\mathcal{Q}$. Taking $\mathcal{P} = (\mathbf{P}, \mathbf{N}, \mathbf{X})$ as an example, it consists of $\mathbf{P}\in \mathbb{R}^{n\times3}$ the points cloud, $\mathbf{N}\in \mathbb{R}^{n \times 3}$ the normals estimated from $\mathbf{P}$, and $\mathbf{X} = \vec{\mathbf{1}}\in\mathbb{R}^{n\times 1}$ the initial point features. The encoder produces the superpoint triplet  $\mathcal{P}^{\prime}=(\mathbf{P}^\prime, \mathbf{N}^\prime, \mathbf{X}^\prime)$ with $\mathbf{P}^\prime\in \mathbb{R}^{n^\prime\times 3}$ and $\mathbf{X}^\prime\in\mathbb{R}^{n^\prime\times c^\prime}$.  With the consecutive decoder, $\mathcal{P}^\prime$ is decoded to a triplet $\hat{\mathcal{P}} = (\hat{\mathbf{P}}, \hat{\mathbf{N}}, \hat{\mathbf{X}})$ including $\hat{n}$ points with features $\hat{\mathbf{X}}\in \mathbb{R}^{\hat{n}\times\hat{c}}$. Notably, as we adopt a Farthest Point Sampling~(FPS) strategy~\cite{qi2017pointnet++}, it always satisfies that $\mathbf{P}^\prime \subseteq \hat{\mathbf{P}} \subseteq \mathbf{P}$.  The same goes for a second point cloud $\mathbf{Q}$ with an input triplet $\mathcal{Q}=(\mathbf{Q}\in \mathbb{R}^{m \times 3}, \mathbf{M}\in \mathbb{R}^{m \times 3}, \mathbf{Y} = \vec{\mathbf{1}} \in \mathbb{R}^{m \times 1})$ by the shared architecture. In the rest of this paper, we only demonstrate for $\mathcal{P}$ unless the model processes $\mathcal{Q}$ differently.

\noindent\textbf{Encoder.} 
The encoder is constructed by stacking several encoder blocks, each including an Attentional Abstraction Layer~(AAL) followed by $e\times$PPF Attention Layers~(PALs). Each block consumes the output of the previous block as the \textit{Support} triplet $\mathcal{P}^{S}$~($\mathcal{P}^{S} = \mathcal{P}$ for the first block). $\mathcal{P}^{S}$ first flows to AAL, where \textit{Anchor} points $\mathbf{P}^{A}$ with associated normals $\mathbf{N}^{A}$ are obtained via FPS~\cite{qi2017pointnet++}. The \textit{Anchor} triplet $\mathcal{P}^{A}$ is then generated in AAL via a PAM following Eq.~\ref{eq:pam}. A sequence of PALs is applied for enhancing the \textit{Anchor} features $\mathbf{X}^{A}$, each updating the features as:
\begin{equation}
    \mathcal{P}^{A} \leftarrow \theta(\mathcal{P}^{A})=\text{ReLU}(\mathbf{X}^{A} + \phi(\delta(\mathbf{P}^{A},\mathbf{N}^{A} \big| \mathcal{P}^{A}))),
\label{eq:PAL}
\end{equation}

\noindent with $\phi$ the LayerNorm~\cite{ba2016layer}, $\delta$ the PAM, and $\theta$ the PAL. $\leftarrow$ depicts feature updating. The encoder block outputs the updated $\mathcal{P}^A$, and the output of the whole encoder is defined as $\mathcal{P}^\prime$, which is the output of the final encoder block.

\noindent\textbf{Decoder.} 
We build the decoder by stacking a series of decoder blocks, each consisting of a Transition Up Layer~(TUL) followed by $d\times$ PAL. Each block takes the output of the previous block as the \textit{Anchor} triplet $\mathcal{P}^{A}$~($\mathcal{P}^{A} = \mathcal{P}^\prime$ for the first block), and takes the \textit{Support} triplet $\mathcal{P}^{S}$ from the encoder via skip connections. The input flows to TUL, where each feature $\widetilde{\mathbf{x}}^{S}_j \in \widetilde{\mathbf{X}}^{S}$ assigned to $\mathbf{p}^S_j\in \mathbf{P}^S$ is interpolated by:
\vspace{-0.3cm}
\begin{equation}
\widetilde{\mathbf{x}}^{S}_j = \frac{\sum_{i\in \mathcal{N}(j)} w_i^j \mathbf{x}^{A}_i}{\sum_{i\in \mathcal{N}(j)} w_i^j} , \ \text{with} \ w_i^j = \frac{1}{\lVert \mathbf{p}^{S}_j - \mathbf{p}^{A}_i\rVert_2},
\label{eq:interpolation}
\end{equation}
\noindent with $\mathcal{N}(j)$ the $k$-nearest neighbors of $\mathbf{p}^S_j$ in $\mathbf{P}^A$. Features are updated by two linear layers as $\mathcal{P}^{S}\leftarrow \zeta_1(\mathbf{X}^{S}) + \zeta_2(\widetilde{\mathbf{X}}^{S})$. A sequence of PALs is adopted after TUL, each enhancing the features as $\mathcal{P}^{S} \leftarrow \theta(\mathcal{P}^{S})$ according to Eq.~\ref{eq:PAL}. The decoder block outputs the updated $\mathcal{P}^S$, and the output of the whole decoder is denoted as $\hat{\mathcal{P}}$, which is the output of the final decoder block.


\begin{figure}
  \includegraphics[width=0.48\textwidth]{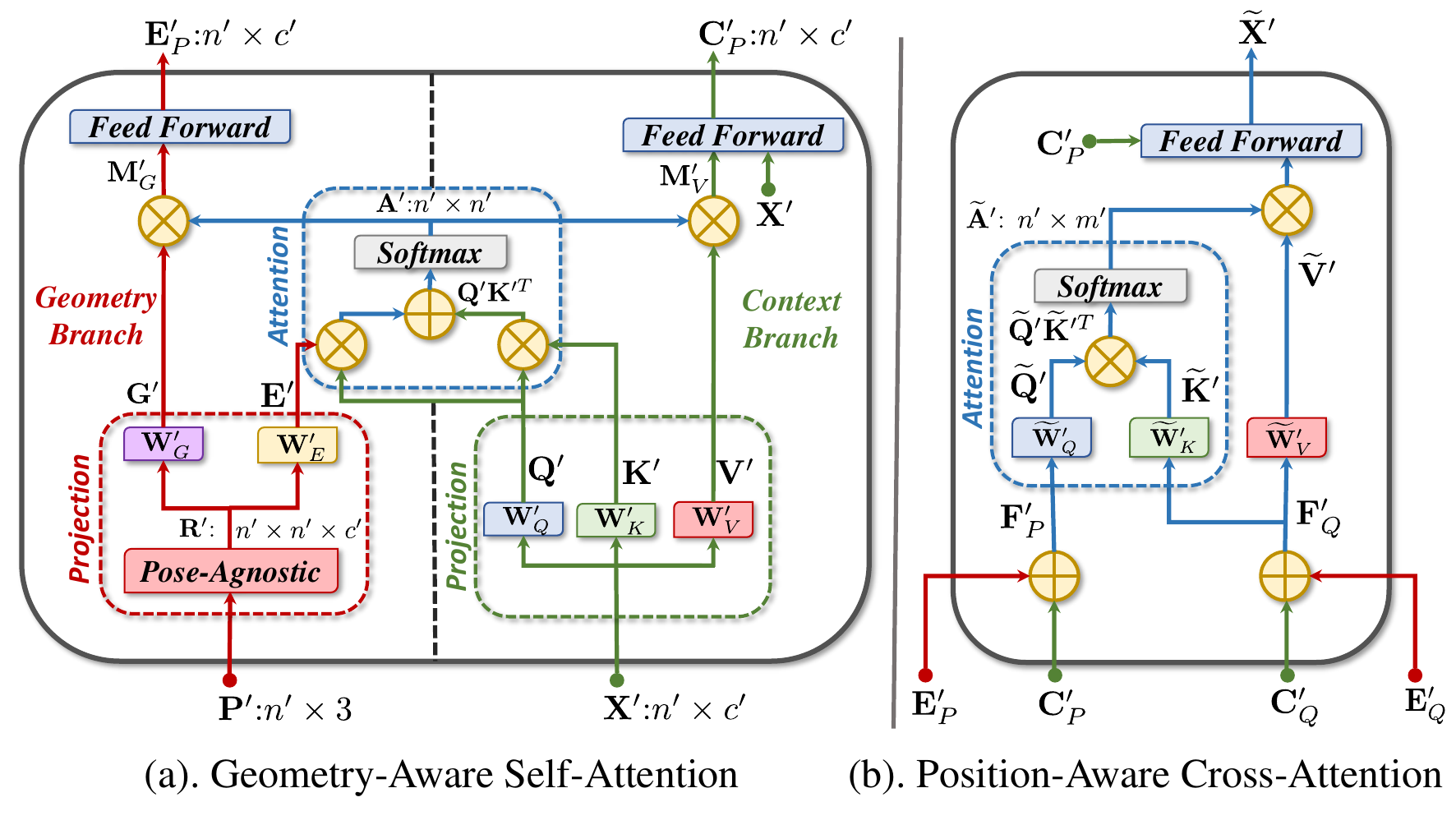}
  \vspace{-0.8cm}
  \caption{The computation graph of our global transformer consisting of the Geometry-Aware Self-Attention Module~(GSM) and Position-Aware Cross-Attention Module~(PCM).}
  \label{fig:global_self}
  \vspace{-0.5cm}
\end{figure}

\subsection{Global Transformer for Context Aggregation}
\label{sec:global_context}
\noindent\textbf{Overview.} Our designed global transformer takes as input a pair of triplets $\mathcal{P}^\prime$ and $\mathcal{Q}^\prime$, and enhances the features with the global context, yielding $\widetilde{\mathbf{X}}^{\prime}\in \mathbb{R}^{n^\prime \times {c}^\prime}$ and $\widetilde{\mathbf{Y}}^{\prime}\in \mathbb{R}^{m^\prime \times {c}^\prime}$, respectively. We stack $g\times$global transformers, with each including a Geometry-Aware Self-Attention Module~(GSM) and a Position-Aware Cross-Attention Module~(PCM)~(See Fig.~\ref{fig:pipeline} and Fig.~\ref{fig:global_self}).
Different from previous works~\cite{huang2021predator,yu2021cofinet,qin2022geometric,yew2022regtr,yu2022riga} that totally neglect the cross-frame spatial relationships, we propose to learn a rotation-invariant position representation for each superpoint to enable the position-aware cross-frame context aggregation.

\noindent\textbf{Geometry-Aware Self-Attention Module.} 
On the global level, we modify PAM to learn the rotation-invariant position representation and to aggregate the learned context across the whole frame simultaneously. The design of GSM is detailed in Fig.~\ref{fig:global_self}~(a). GSM has two branches, where the geometry branch mines the geometric cues from the pairwise rotation-invariant geometry representation proposed in~\cite{qin2022geometric}, and the context branch aggregates the global context across the frame. We refer the readers to the Appendix for the detailed construction of $\mathbf{R}^\prime\in \mathbb{R}^{n^\prime \times n^\prime \times c^\prime}$ and the ablation study on it. Similar to Eq.~\ref{eq:project1}, the geometric cues $\mathbf{G}^\prime$ and the positional encoding $\mathbf{E}^\prime$ are linearly projected from $\mathbf{R}^\prime$. $\mathbf{E}^\prime$ is further processed in the geometry branch and finally leveraged as the rotation-invariant position representation. In the context branch, $\mathbf{Q}^\prime$, $\mathbf{K}^\prime$, and $\mathbf{V}^\prime$ are obtained by linearly mapping the input features $\mathbf{X}^\prime$ similar to Eq.~\ref{eq:project2}. The hybrid score matrix $\mathbf{S}^\prime\in \mathbb{R}^{n^\prime \times n^\prime}$ is computed as:

\begin{equation}
    \mathbf{S}^\prime(i, j) = \frac{(\mathbf{q}^\prime_i)(\mathbf{e}^\prime_{i, j} + \mathbf{k}^\prime_j)^T}{\sqrt{c^\prime}},
\label{eq:hybrid_score}
\end{equation}
with $\mathbf{e}^\prime_{i, j} := \mathbf{E}^\prime(i, j, :)$, $\mathbf{q}^\prime_{i} := \mathbf{Q}^\prime(i, :)$, and $\mathbf{k}^\prime_{j} := \mathbf{K}^\prime(j, :)$ the $c^\prime$-dimension vectors.
The hybrid attention $\mathbf{A}^\prime$ is obtained via a Softmax function over each row of $\mathbf{S}^\prime$, and the geometric messages $\mathbf{M}^\prime_G \in \mathbb{R}^{n^\prime \times c^\prime}$ are computed as:

\begin{equation}
\mathbf{M}_G^\prime(i, :) = \sum_{1\leq j \leq n^\prime} a^\prime_{i, j}\mathbf{g}^\prime_{i, j},
\label{eq:geo_mes}
\end{equation}

\noindent with $a^\prime_{i, j} := \mathbf{A}^\prime(i, j)$ and $\mathbf{g}^\prime_{i, j} := \mathbf{G}^\prime(i, j, :)$. The contextual messages $\mathbf{M}_V^\prime \in \mathbb{R}^{n^\prime\times c^\prime}$ are computed by $\mathbf{A}^\prime\mathbf{V}^\prime$. After a feed-forward network~\cite{vaswani2017attention}, the position representation $\mathbf{E}^\prime_P$ and globally-enhanced context $\mathbf{C}^\prime_P$ are generated.

\noindent\textbf{Position-Aware Cross-Attention Module.} PCM consumes a pair of doublets $(\mathbf{E}^\prime_P, \mathbf{C}^\prime_P)$ and $(\mathbf{E}^\prime_Q, \mathbf{C}^\prime_Q)$ that are generated from $\mathcal{P}^\prime$ and $\mathcal{Q}^\prime$ by a shared GSM, respectively. As the cross-attention is directional, we apply the same PCM twice, with the first aggregation from $\mathcal{Q}^\prime$ to $\mathcal{P}^\prime$~(See Fig.~\ref{fig:global_self} (b)), and the second reversed. As the first step, the rotation-invariant position representation is incorporated to make the consecutive cross-attention position-aware, yielding position-aware features $\mathbf{F}^\prime_P=\mathbf{E}^\prime_P + \mathbf{C}^\prime_P$ and $\mathbf{F}^\prime_Q=\mathbf{E}^\prime_Q + \mathbf{C}^\prime_Q$. Similar to Eq.~\ref{eq:project2}, $\widetilde{\mathbf{Q}}^\prime$, $\widetilde{\mathbf{K}}^\prime$, and $\widetilde{\mathbf{V}}^\prime$ are computed as the linear projection of $\mathbf{F}^\prime_P$, $\mathbf{F}^\prime_Q$, and $\mathbf{F}^\prime_Q$, respectively. The attention matrix $\widetilde{\mathbf{A}}\in \mathbb{R}^{n^\prime \times m^\prime}$ is computed via a row-wise softmax function applied on $\widetilde{\mathbf{Q}}^\prime\widetilde{\mathbf{K}}^{\prime T}$. The fused messages are presented as $\widetilde{\mathbf{A}}\widetilde{\mathbf{V}}^\prime$, which are finally mapped to the output features $\widetilde{\mathbf{X}}^\prime$ through a feed-forward network. As we introduce spatial awareness at the beginning of PCM, both the attention computation and message fusion are aware of the cross-frame positions. After the twice application of PCM, the input features are enhanced as $\mathcal{P}^\prime\leftarrow\widetilde{\mathbf{X}}^\prime$ and $\mathcal{Q}^\prime\leftarrow\widetilde{\mathbf{Y}}^\prime$, respectively. The global aggregation stage finally generates a pair of triplets $\widetilde{\mathcal{P}}^\prime:=(\mathbf{P}^\prime, \mathbf{N}^\prime, \widetilde{\mathbf{X}}^\prime)$ and $\widetilde{\mathcal{Q}}^\prime:=(\mathbf{Q}^\prime, \mathbf{M}^\prime, \widetilde{\mathbf{Y}}^\prime)$, with the enhanced features from the last global transformer.

\subsection{Point Matching and Loss Funcion}

\label{sec:matching}
\noindent\textbf{Superpoint Matching.} As shown in Fig.~\ref{fig:pipeline}, the point matching stage consumes a pair of superpoint triplets $\widetilde{\mathcal{P}}^\prime$ and $\widetilde{\mathcal{Q}}^\prime$ obtained from the global transformer, as well as a pair of point triplets $\hat{\mathcal{P}}$ and $\hat{\mathcal{Q}}$ produced by the decoder. 
We adopt the coarse-to-fine matching proposed in~\cite{yu2021cofinet}. Following~\cite{qin2022geometric}, we first normalize the superpoint features $\widetilde{\mathbf{X}}^\prime$ and $\widetilde{\mathbf{Y}}^{\prime}$ onto a unit hypersphere, and measure the pairwise similarity using a Gaussian correlation matrix $\widetilde{\mathbf{S}}$ with $\widetilde{\mathbf{S}}(i, j)=-\text{exp}(-\lVert\widetilde{\mathbf{x}}_i^\prime - \widetilde{\mathbf{y}}_j^\prime\rVert_2^2)$. After a dual-normalization~\cite{rocco2018neighbourhood,sun2021loftr,qin2022geometric} on $\widetilde{\mathbf{S}}$ for global feature correlation, superpoints associated to the top-$k$ entries are selected as the coarse correspondence set $\mathcal{C}^\prime = \{(\mathbf{p}_i^\prime, \mathbf{q}_j^\prime)\big| \mathbf{p}_i^\prime \in \mathbf{P}^\prime, \mathbf{q}_j^\prime \in \mathbf{Q}^\prime\}$.

\noindent\textbf{Point Matching.} For extracting point correspondences, denser points $\hat{\mathbf{P}}$ and $\hat{\mathbf{Q}}$ are first assigned to superpoints. To this end, the point-to-node strategy~\cite{yu2021cofinet} is leveraged, where each point is assigned to its closest superpoint in 3D space. Given a superpoint $\mathbf{p}_i^\prime \in \mathbf{P}^\prime$, the group of points assigned to it is denoted as $\hat{\mathbf{G}}^P_i \subseteq \hat{\mathbf{P}}$. The group of features associated to $\hat{\mathbf{G}}^P_i$ is further defined as $\hat{\mathbf{G}}^X_i$ with $\hat{\mathbf{G}}^X_i \subseteq \hat{X}$. For each superpoint correspondence $\mathcal{C}^\prime_l = (\mathbf{p}^\prime_i, \mathbf{q}^\prime_j)$, the similarity between the corresponding feature groups $\hat{\mathbf{G}}^X_i$ and $\hat{\mathbf{G}}^Y_j$ is calculated as
$\hat{\mathbf{S}}_l = \hat{\mathbf{G}}^X_i (\hat{\mathbf{G}}_j^{Y})^T /\sqrt{\hat{c}}$, with $\hat{c}$ the feature dimension. We then follow~\cite{sarlin2020superglue} to append a stack row and column to $\hat{\mathbf{S}}_l$ filled with a learnable parameter $\alpha$, and iteratively run the Sinkhorn Algorithm~\cite{sinkhorn1967concerning}. After removing the slack row and column of $\hat{\mathbf{S}}_l$, the mutual top-k entries, \ie., entries with top-$k$ confidence on both the row and the column, are selected to formulate a point correspondence set $\hat{\mathcal{C}}_l$. The final correspondence set $\hat{\mathcal{C}}$ is collected by $\hat{\mathcal{C}} = \cup^{|\mathcal{C}^\prime|}_{l=1} \hat{\mathcal{C}}_l$. 

\noindent\textbf{Loss Function.} Our loss function reads as $\mathcal{L} = \mathcal{L}_s + \lambda\mathcal{L}_p$, with a superpoint matching loss $\mathcal{L}_s$ and a point matching loss $\mathcal{L}_p$ balanced by a hyper-parameter $\lambda$~($\lambda=1$ by default). The detailed definition is introduced in the Appendix.

\section{Experiment}
\label{sec:experiment}
We evaluate \OURS{} on both rigid~(3DMatch~\cite{zeng20173dmatch} \& 3DLoMatch~\cite{huang2021predator}) and non-rigid~(4DMatch~\cite{li2022lepard} \& 4DLoMatch~\cite{li2022lepard}) benchmarks. For the rigid matching, we further evaluate our correspondences on the registration task, where RANSAC~\cite{fischler1981random} is used. Details of implementation, metrics, and runtime analysis are introduced in the Appendix.

\renewcommand\arraystretch{0.75}
\begin{table}[ht!]
\small
\centering
\resizebox{0.48\textwidth}{!}{
\begin{tabular}{lcc|cc}
\toprule
&\multicolumn{2}{c}{\textbf{3DMatch}} &\multicolumn{2}{c}{\textbf{3DLoMatch}}\\
\# Samples=5,000 &Origin &Rotated &Origin &Rotated\\
\midrule
\midrule

&\multicolumn{4}{c}{\textit{Feature Matching Recall} (\%) $\uparrow$} \\
\midrule
SpinNet~\cite{ao2021spinnet}  &97.4 &{97.4}  &75.5 &75.2 \\
Predator~\cite{huang2021predator}  &96.6 &96.2 &78.6  &73.7 \\
CoFiNet~\cite{yu2021cofinet}  &\underline{98.1} &{97.4} &{83.1} &{78.6}\\
YOHO~\cite{wang2022you} &\textbf{98.2} &{97.8} &79.4 &77.8\\
RIGA\cite{yu2022riga}~ &97.9 &\textbf{98.2}  &{85.1} &{84.5}\\
Lepard~\cite{li2022lepard} &98.0 &97.4 &{83.1} &{79.5}\\
GeoTrans~\cite{qin2022geometric} &97.9 &97.8 &\underline{88.3} &\underline{85.8}\\
\OURS{}~(\textit{Ours}) &98.0 &\textbf{98.2} &\textbf{89.6} &\textbf{89.4} \\
\midrule

&\multicolumn{4}{c}{\textit{Inlier Ratio} (\%) $\uparrow$}\\
\midrule
SpinNet~\cite{ao2021spinnet} &48.5 &48.7 &25.7 &{25.7}\\
Predator~\cite{huang2021predator} &58.0 &{52.8}  &26.7 &22.4\\
CoFiNet~\cite{yu2021cofinet} &49.8 &46.8 &24.4  &21.5\\
YOHO~\cite{wang2022you} &{64.4} &{64.1} &25.9 &23.2\\
RIGA~\cite{yu2022riga}  &{68.4} &\underline{68.5} &{32.1} &{32.1}\\
Lepard~\cite{li2022lepard} &58.6 &53.7 &{28.4} &{24.4}\\
GeoTrans~\cite{qin2022geometric} &\underline{71.9} &68.2 &\underline{43.5} &\underline{40.0}\\
\OURS{}~(\textit{Ours}) &\textbf{82.6} &\textbf{82.3} &\textbf{54.3} &\textbf{53.2} \\
\midrule

&\multicolumn{4}{c}{\textit{Registration Recall} (\%) $\uparrow$} \\
\midrule
SpinNet~\cite{ao2021spinnet}  &88.8 &\underline{93.2}  &58.2 &61.8\\
Predator~\cite{huang2021predator}  &{89.0} &92.0  &59.8 &58.6\\
CoFiNet~\cite{yu2021cofinet}  &{89.3} &92.0  &{67.5} &{62.5}\\
YOHO~\cite{wang2022you} &{90.8} &92.5 &65.2 &{66.8}\\
RIGA~\cite{yu2022riga}~ &{89.3} &{93.0}  &{65.1} &{66.9}\\
Lepard~\cite{li2022lepard} &\textbf{92.7} &84.9 &{65.4} &49.0\\
GeoTrans~\cite{qin2022geometric} &\underline{92.0} &92.0 &\textbf{75.0} &\underline{71.8}\\
\OURS{}~(\textit{Ours}) &91.9 &\textbf{94.7} &\underline{74.8} &\textbf{77.2}\\
\bottomrule
\end{tabular}}
\vspace{-0.25cm}
\caption{Quantitative results on (Rotated) 3DMatch \& 3DLoMatch. 5,000 points/correspondences are used for the evaluation.}
\label{tab:3dmatch}
\vspace{-0.65cm}
\end{table}

\begin{figure*}
  \includegraphics[width=0.98\textwidth]{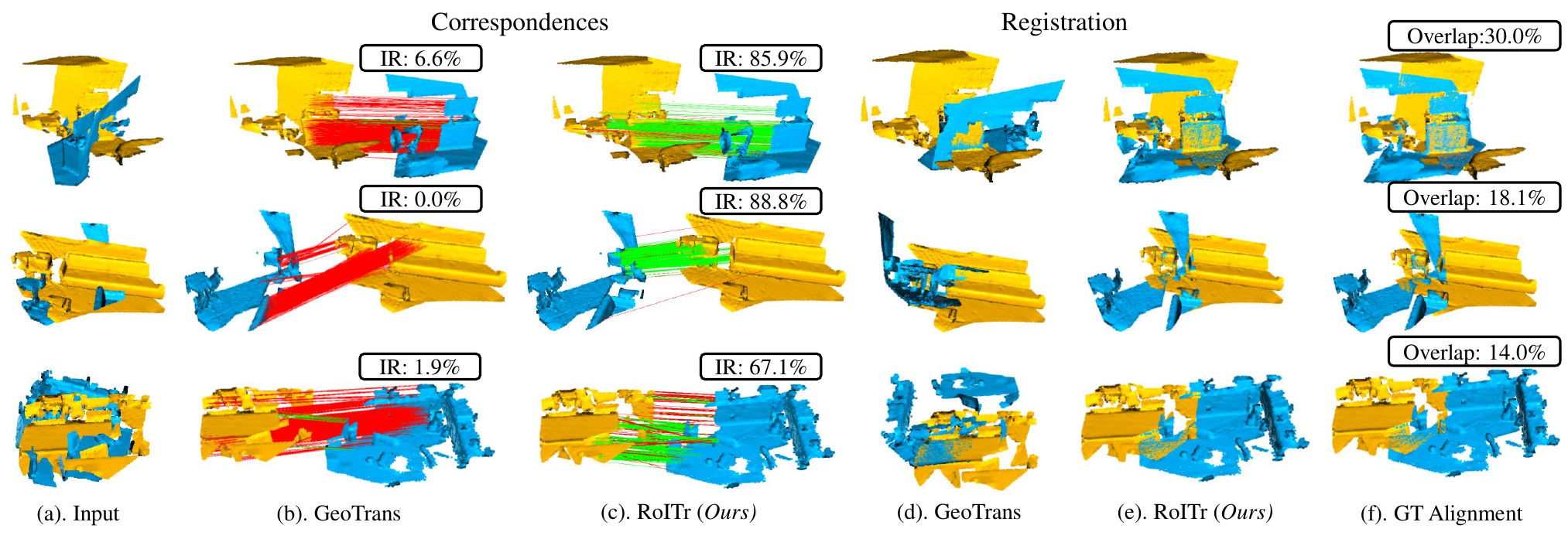}
  \vspace{-0.3cm}
  \caption{Qualitative results on 3DLoMatch. GeoTrans~\cite{qin2022geometric} is used as the baseline. Columns (b) and (c) show the correspondences, while columns (d) and (e) demonstrate the registration results. Green/red lines indicate inliers/outliers. More cases are shown in the Appendix.}
  \label{fig:vis_indoor}
  \vspace{-0.55cm}
\end{figure*}

\subsection{Rigid Indoor Scenes: 3DMatch \& 3DLoMatch}
\noindent\textbf{Dataset.}
3DMatch~\cite{zeng20173dmatch} collects 62 indoor scenes, among which 46 are used for training, 8 for validation, and 8 for testing. We use the data processed by~\cite{huang2021predator} where the 3DMatch data is spilt as 3DMatch~($>30\%$ overlap) and 3DLoMatch~($10\%\sim 30\%$ overlap). To evaluate robustness to arbitrary rotations, we follow~\cite{yu2022riga} for creating the rotated benchmarks, where full-range rotations are individually added to the two frames of each point cloud pair.

\noindent\textbf{Metrics.} We follow~\cite{huang2021predator} to use three metrics for evaluation: (1).~\textit{Inlier Ratio}~(IR) that computes the ratio of putative correspondences whose residual distance is smaller than a threshold~(\ie., 0.1m) under the ground-truth transformation; (2).~\textit{Feature Matching Recall}~(FMR) that calculates the fraction of point cloud pairs whose IR is larger than a threshold~(\ie., 5\%); (3).~\textit{Registration Recall}~(RR) that counts the fraction of point cloud pairs that are correctly registered~(\ie., with RMSE $<$ 0.2m).~\footnote{We follow~\cite{yu2022riga} to calculate the RR strictly with RMSE $<$ 0.2m on the rotated data, which is slightly different from the RR on the original data.}

\noindent\textbf{Comparison with the State-of-the-Art.}
We compare \OURS{} with 7 state-of-the-art methods, among which Predator~\cite{huang2021predator}, CoFiNet~\cite{yu2021cofinet}, Lepard~\cite{li2022lepard}, and GeoTrans~\cite{qin2022geometric} are rotation-sensitive models, while SpinNet~\cite{ao2021spinnet}, YOHO~\cite{wang2022you}, and RIGA~\cite{yu2022riga} guarantee the rotation invariance by design. In Tab.~\ref{tab:3dmatch} we demonstrate the matching and registration results on 3DMatch and 3DLoMatch, as well as on their rotated versions, with 5,000 sampled points/correspondences. Regarding IR, \OURS{} outperforms all the others by a large margin on both datasets, which indicates our method matches points more correctly. For FMR, we significantly surpass all the others on 3DLoMatch, while staying on par with CoFiNet and YOHO on 3DMatch, which indicates that our model is good at coping with hard cases, \ie., we find at least 5\% inliers on more test data. For the registration evaluation in terms of RR, \OURS{} achieves comparable performance with GeoTrans and Lepard on 3DMatch, but leads the board together with GeoTrans on 3DLoMatch with an overwhelming advantage over the others. Our stability against additional rotations is further demonstrated on the rotated data, where we outperform all the others with a substantial margin. Qualitative results can be found in Fig.~\ref{fig:vis_indoor}.

\renewcommand\arraystretch{0.95}
\begin{table}[ht!]
\centering

\resizebox{0.48\textwidth}{!}{
\begin{tabular}{lccccc|ccccc}
\toprule
 &\multicolumn{5}{c}{\textbf{3DMatch}}  &\multicolumn{5}{c}{\textbf{3DLoMatch}}\\
\# Samples &5000 &2500 &1000 &500 &250 &5000 &2500 &1000 &500 &250 \\
\midrule
\midrule
&\multicolumn{10}{c}{\textit{Feature Matching Recall} (\%) $\uparrow$}\\
\midrule
SpinNet~\cite{ao2021spinnet} &97.4 &97.0 &96.4 &96.7 &94.8 &75.5 &75.1 &74.2 &69.0 &62.7\\
Predator~\cite{huang2021predator} &96.6 &96.6 &96.5 &96.3 &96.5 &{78.6} &{77.4} &{76.3} &{75.7} &{75.3}\\
CoFiNet~\cite{yu2021cofinet} &\underline{98.1} &\textbf{98.3} &\textbf{98.1} &\textbf{98.2} &\textbf{98.3} &{83.1} &{83.5} &{83.3} &{83.1} &{82.6}\\

YOHO~\cite{wang2022you} &\textbf{98.2} &97.6 &97.5 &{97.7} &96.0 &79.4 &78.1 &76.3 &73.8 &69.1\\
RIGA~\cite{yu2022riga} &97.9 &{97.8} &{97.7} &{97.7} &{97.6} &{85.1} &{85.0} &{85.1} &{84.3} &{85.1}\\
GeoTrans~\cite{qin2022geometric} &97.9 &97.9 &\underline{97.9} &97.9 &97.6 &\underline{88.3} &\underline{88.6} &\underline{88.8} &\underline{88.6} &\underline{88.3}\\
\OURS{}~(\textit{Ours}) &98.0 &\underline{98.0} &\underline{97.9} &\underline{98.0} &\underline{97.9} &\textbf{89.6} &\textbf{89.6} &\textbf{89.5} &\textbf{89.4} &\textbf{89.3}\\
\midrule
&\multicolumn{10}{c}{\textit{Inlier Ratio} (\%) $\uparrow$}\\
\midrule
SpinNet~\cite{ao2021spinnet} &48.5 &46.2 &40.8 &35.1 &29.0 &25.7 &23.7 &20.6 &18.2 &13.1\\
Predator~\cite{huang2021predator} &58.0 &58.4 &{57.1} &{54.1} &{49.3}  &{26.7} &{28.1} &{28.3} &{27.5} &{25.8}\\
CoFiNet~\cite{yu2021cofinet} &49.8 &51.2 &{51.9} &{52.2} &{52.2} &{24.4} &{25.9} &{26.7} &{26.8} &{26.9}\\
YOHO~\cite{wang2022you} &{64.4} &{60.7} &55.7 &46.4 &41.2 &25.9 &23.3 &22.6 &18.2 &15.0 \\
RIGA~\cite{yu2022riga}  &{68.4} &{69.7} &{70.6} &{70.9} &{71.0} &{32.1}  &{33.4} &{34.3} &{34.5} &{34.6}\\
GeoTrans~\cite{qin2022geometric} &\underline{71.9} &\underline{75.2} &\underline{76.0} &\underline{82.2} &\textbf{85.1} &\underline{43.5} &\underline{45.3} &\underline{46.2} &\underline{52.9} &\textbf{57.7}\\
\OURS{}~(\textit{Ours}) &\textbf{82.6} &\textbf{82.8} &\textbf{83.0} &\textbf{83.0} &\underline{83.0} &\textbf{54.3} &\textbf{54.6} &\textbf{55.1} &\textbf{55.2} &\underline{55.3}\\
\midrule

&\multicolumn{10}{c}{\textit{Registration Recall} (\%) $\uparrow$}\\
\midrule
SpinNet~\cite{ao2021spinnet} &88.8 &88.0 &84.5 &79.0 &69.2 &58.2 &56.7 &49.8 &41.0 &26.7\\
Predator~\cite{huang2021predator}&89.0 &{89.9} &{90.6} &{88.5} &{86.6} &{59.8} &{61.2} &{62.4} &{60.8} &{58.1}\\
CoFiNet~\cite{yu2021cofinet} &{89.3} &{88.9} &{88.4} &{87.4} &{87.0} &{67.5} &{66.2} &{64.2} &{63.1} &{61.0}\\

YOHO~\cite{wang2022you}&{90.8} &{90.3} &{89.1} &{88.6} &84.5 &{65.2} &{65.5} &63.2 &56.5 &48.0\\

RIGA~\cite{yu2022riga} &{89.3} &88.4 &{89.1} &{89.0} &{87.7}&{65.1} &{64.7} &{64.5} &{64.1} &{61.8}\\
GeoTrans~\cite{qin2022geometric} &\textbf{92.0} &\textbf{91.8} &\textbf{91.8} &\textbf{91.4} &\textbf{91.2} &\textbf{75.0} &\textbf{74.8} &\underline{74.2} &\underline{74.1} &\underline{73.5}\\
\OURS{}~(\textit{Ours}) &\underline{91.9} &\underline{91.7} &\textbf{91.8} &\textbf{91.4} &\underline{91.0} &\underline{74.7} &\textbf{74.8} &\textbf{74.8} &\textbf{74.2} &\textbf{73.6}\\
\bottomrule
\end{tabular}}
\vspace{-0.2cm}
\caption{Quantitative results on 3DMatch \& 3DLoMatch with a varying number of points/correspondences. See the results on rotated data in the Appendix.}
\label{tab:scene}
\vspace{-0.55cm}

\end{table}

\noindent\textbf{Analysis on the Number of Correspondences.} We further analyze the influence of a varying number of correspondences. As illustrated in Tab.~\ref{tab:scene}, \OURS{} shows outstanding performance on both datasets with various correspondences, proving its stability when only a few correspondences are accessible. The same test on the rotated benchmarks is given in the Appendix.

\subsection{Deformable Objects: 4DMatch \& 4DLoMatch}

\noindent\textbf{Dataset.} 4DMatch~\cite{li2022lepard} contains 1,761 animations randomly selected from DeformingThings4D~\cite{li20214dcomplete}. The 1,761 sequences are divided into 1,232/176/353 as train/val/test, where the test set is further split into 4DMatch and 4DLoMatch based on an overlap ratio threshold of 45$\%$.
\renewcommand\arraystretch{0.9}
\begin{table}[ht!]
\centering
\resizebox{0.48\textwidth}{!}{
\begin{tabular}{c|l|cc|cc}
\toprule
& &\multicolumn{2}{c}{\textbf{4DMatch}} &\multicolumn{2}{c}{\textbf{4DLoMatch}}\\
Category & Method &NFMR(\%) $\uparrow$ &IR(\%) $\uparrow$ &NFMR(\%) $\uparrow$ &IR(\%) $\uparrow$\\
\midrule
\midrule
\multirow{2}{*}{Scene Flow}&PWC~\cite{wu2019pointpwc}  &21.6 &20.0 &10.0 &7.2\\
&FLOT~\cite{puy2020flot} &27.1 &24.9 &15.2 &10.7\\

\midrule
\multirow{4}{*}{Feature Matching}
&Predator~\cite{huang2021predator} &{56.4} &{60.4} &32.1 &27.5\\
&GeoTrans~\cite{qin2022geometric} &\underline{83.2} &82.2 &65.4 &\underline{63.6}\\
&Lepard~\cite{li2022lepard} &\textbf{83.7} &\underline{82.7} &\underline{66.9} &{55.7}\\
&\OURS{}~(\textit{Ours}) &{83.0} &\textbf{84.4} &\textbf{69.4} &\textbf{67.6} \\
\bottomrule
\end{tabular}}
\vspace{-0.2cm}
\caption{Quantitative results on 4DMatch \& 4DLoMatch.}

\label{tab:4dmatch}
\vspace{-0.2cm}
\end{table}
\noindent\textbf{Metrics.}
We follow~\cite{li2022lepard} to use two different metrics: (1).~\textit{Inlier Ratio}~(IR) which is defined as same as the IR on 3DMatch, but with a different threshold~(\ie., 0.04m); (2). \textit{Non-rigid Feature Matching Recall}~(NFMR) that measures the fraction of ground-truth matches that can be successfully recovered by the putative correspondences.

\begin{figure}
\includegraphics[width=0.48\textwidth]{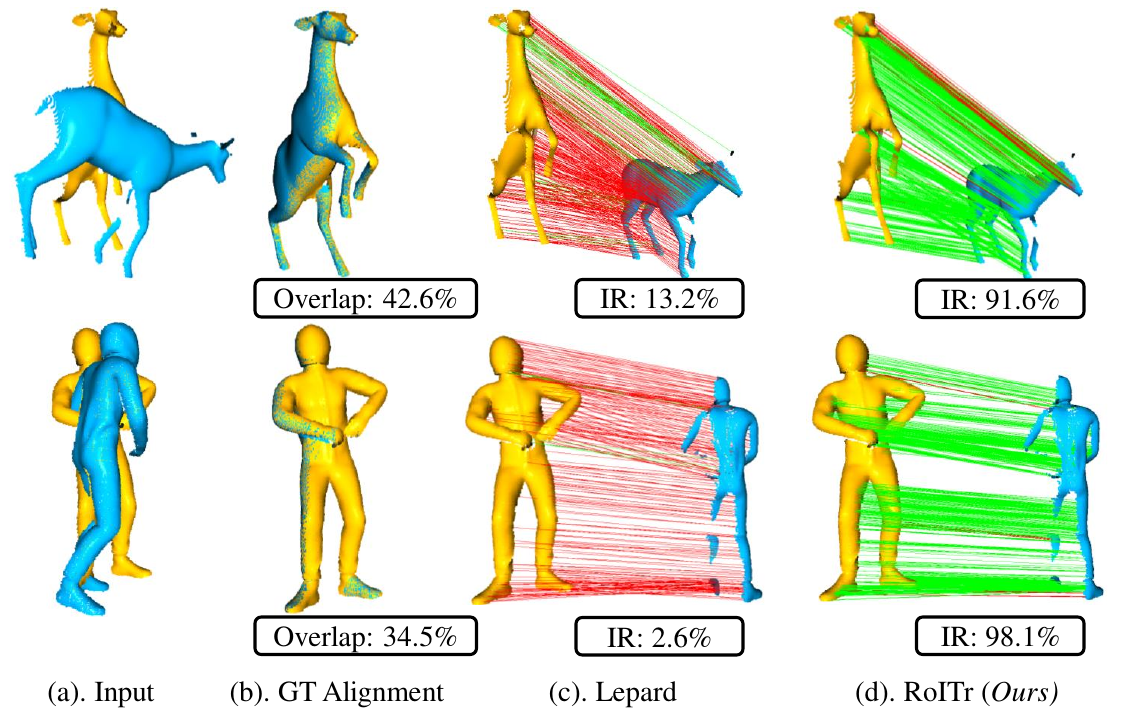}
\vspace{-0.5cm}
\caption{\small Qualitative results of non-rigid matching on 4DLoMatch with Lepard~\cite{li2022lepard} as the baseline. Green/red lines indicate inliers/outliers. See the Appendix for more examples.}
\label{fig:vis_4d}
\vspace{-0.7cm}
\end{figure}

\noindent\textbf{Comparison with the State-of-the-Art.} We compare \OURS{} with 5 baselines, among which PWC~\cite{wu2019pointpwc} and FLOT~\cite{puy2020flot} are scene flow-based methods, while Predator~\cite{huang2021predator}, Lepard~\cite{li2022lepard}, GeoTrans~\cite{qin2022geometric} are based on feature matching. The results shown in Tab.~\ref{tab:4dmatch} indicate that although our rotation-invariance is mainly designed for rigid scenarios, \OURS{} could also achieve outstanding performance in the non-rigid matching task, which further confirms the superiority of our model design. Qualitative results are demonstrated in Fig.~\ref{fig:vis_4d}.
\subsection{Ablation Study}

\begin{figure}
  \includegraphics[width=0.46\textwidth]{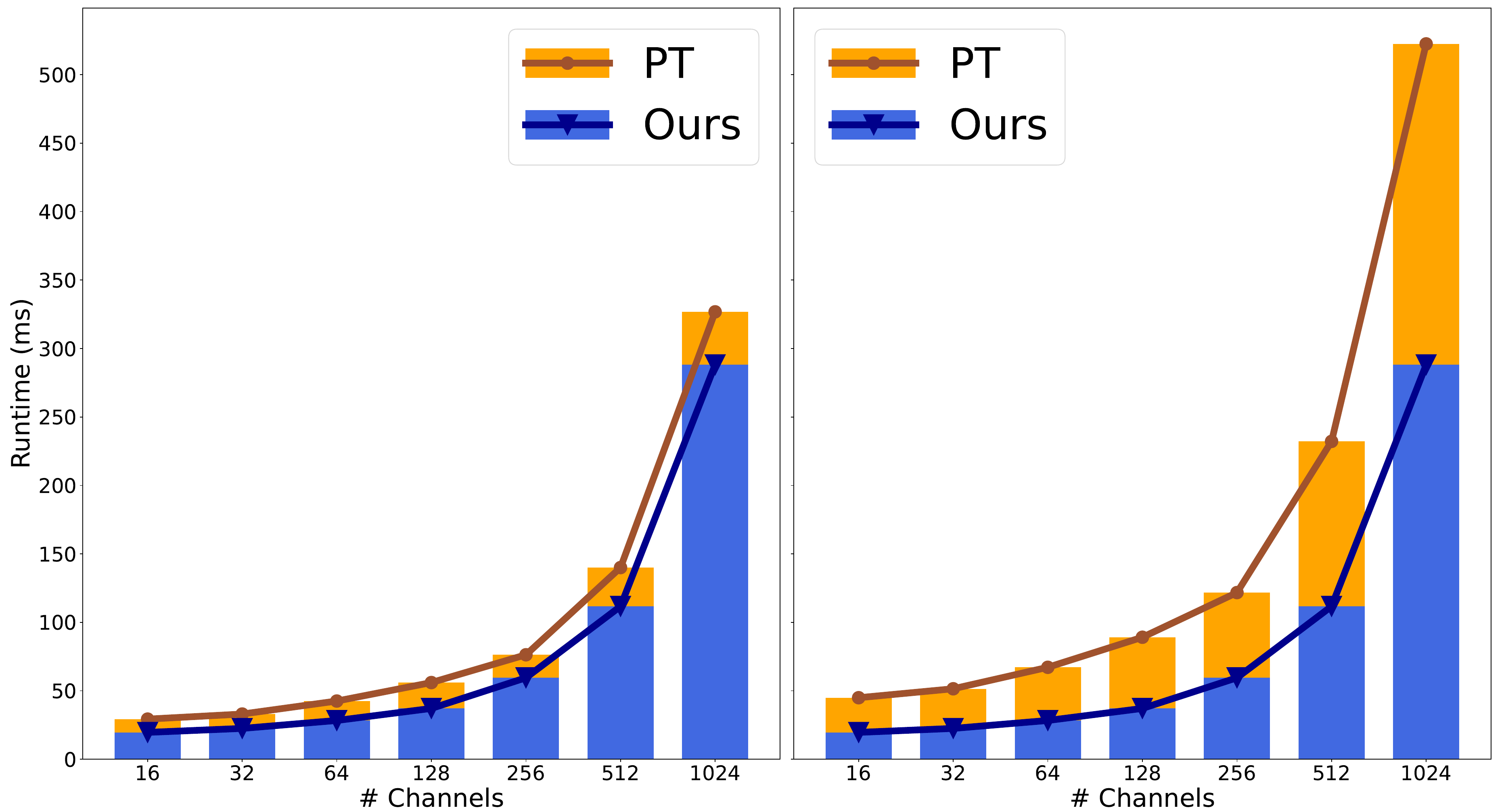}
  \vspace{-0.2cm}
  \caption{\small \textbf{Left}: Runtime comparison between our PPF attention Mechanism~(PAM) and the local attention in PointTransformer~\cite{zhao2021point}. \textbf{Right}: Runtime normalized by aligning the number of parameters. }
  \label{fig:runtime}

\end{figure}

\noindent\textbf{Local Attention.} We first replace our PPFTrans with PointTransformer~(PT)~\cite{zhao2021point} in Tab.~\ref{tab:ablation}~(a.1), which leads to a sharp performance drop. We then ablate by embedding our PPF-based local coordinates into PT~(Tab.~\ref{tab:ablation}~(a.2)) and by adopting the relative coordinates, \ie., $\mathbf{p}_j - \mathbf{p}_j$, used by PT in our PAM~(Tab.~\ref{tab:ablation}~(a.3)). Our local coordinate representation significantly boosts the performance of PT in the task of point cloud matching and meanwhile makes it rotation-invariant, although its performance is still far behind ours. However, the relative coordinates fail to work in our PAM, as we adopt a more efficient attention mechanism~\cite{vaswani2017attention} that learns a scalar attention value for each feature $\mathbf{x}\in\mathbb{R}^c$ and is consequently hard to work under varying poses with a rotation-sensitive design. As a comparison, PT learns a per-channel vector attention $\mathbf{a}\in \mathbb{R}^c$ for the same feature $\mathbf{x}$ and could deal with the pose variations, but at the cost of the efficiency as shown in Fig.~\ref{fig:runtime}. When the number of channels is increased, our advantage in terms of efficiency is enlarged. As we achieve that with more parameters, the gap becomes more significant when runtime is normalized with the number of parameters in the right figure. With our PPF-based local coordinate, the scalar attention could focus on the pose-agnostic pure geometry and therefore achieves the best performance shown in Tab.~\ref{tab:ablation}~(a.4).

\noindent\textbf{Abstraction Layer.}
We ablate our Attentional Abstraction Layer~(AAL) by replacing it with the pooling-based abstraction design used in~\cite{qi2017pointnet,qi2017pointnet++,zhao2021point}. We test the max pooling in Tab.~\ref{tab:ablation}~(b.1) and the average pooling in Tab.~\ref{tab:ablation}~(b.2), both showing a degrading performance compared with our AAL, which demonstrates our superiority.

\noindent \textbf{Backbone.} In Tab.~\ref{tab:ablation}~(a.1) we have shown our superiority compared with PT~\cite{zhao2021point}. We further replace our PPFTrans with the KPConv-based backbone network which is widely used in previous deep matchers~\cite{huang2021predator,yu2021cofinet,qin2022geometric}. 
The fact that KPConv falls behind our design demonstrates the advantage of PPFTrans in geometry encoding.

\noindent \textbf{Global Transformer.} We replace our design with the global transformer of GeoTrans~\cite{qin2022geometric} which performs state-of-the-art but without the cross-frame spatial awareness. The dropping results in Tab.~\ref{tab:ablation}~(d.1) proves the excellence of our design with the cross-frame position awareness.
\renewcommand\arraystretch{0.8}
\begin{table}[ht!]
\vspace{-0.25cm}
\centering
\resizebox{0.48\textwidth}{!}{
\begin{tabular}{ll|cccccc}
\toprule
& &\multicolumn{3}{c}{\textbf{Origin}}&\multicolumn{3}{c}{\textbf{Rotated}}\\
Category & Model &FMR &IR &RR  &FMR &IR &RR\\
\midrule
\midrule
\multirow{4}{*}{a. Local}
 &\;\ 1. PT~\cite{zhao2021point} &79.0&36.5&61.6 &76.5&34.7&60.0 \\
&*2. PPF+PT~\cite{zhao2021point} &87.0 &49.9 &69.9 &86.8 &49.4 &71.2\\
& \; 3. $\Delta$xyz+Ours &-&-&-&-&-&-\\
& *4. \textit{Ours}  &\textbf{89.6} &\textbf{54.3} &\textbf{74.7} &\textbf{89.4} &\textbf{53.2} &\textbf{77.2}\\
\midrule
\multirow{3}{*}{b. Aggregation}& *1. max pooling &85.2&50.1&70.5&85.4&50.2&71.9\\
& *2. avg pooling  &87.8&52.6&73.8&87.2&52.5&74.7\\
& *3. \textit{Ours} &\textbf{89.6} &\textbf{54.3} &\textbf{74.7} &\textbf{89.4} &\textbf{53.2} &\textbf{77.2}\\
\midrule
\multirow{2}{*}{c. Backbone}&\;\ 1. KPConv~\cite{thomas2019kpconv}  &85.2 &44.4 &70.6 &83.0 &42.3 &71.5 \\
& *2. \textit{Ours}  &\textbf{89.6} &\textbf{54.3} &\textbf{74.7} &\textbf{89.4} &\textbf{53.2} &\textbf{77.2}\\
\midrule
\multirow{2}{*}{d. Global} &*1. GeoTrans~\cite{qin2022geometric}  &87.7&53.6&73.0 &87.5&\textbf{53.2} &75.1\\
&*2. \textit{Ours}  &\textbf{89.6} &\textbf{54.3} &\textbf{74.7} &\textbf{89.4} &\textbf{53.2} &\textbf{77.2}\\
\midrule
\multirow{4}{*}{e. \#Global}
& *1. $g=0$  &87.2 &37.6 &70.7 &87.5 &37.6 &72.7 \\
& *2. $g=1$  &87.1 &42.1 &70.8 &86.8 &42.1 &73.0 \\
& *3. $g=3$~(\textit{Ours}) &\textbf{89.6} &\textbf{54.3} &\textbf{74.7} &\textbf{89.4} &\textbf{53.2} &\textbf{77.2}\\
& *4. $g=5$ &87.1 &52.5 &72.1 &87.0&52.4&73.3\\

\bottomrule
\end{tabular}}
\vspace{-0.2cm}
\caption{\small Ablation study on (rotated) 3DLoMatch. 5,000 points/correspondences are leveraged. * indicates the methods with intrinsic rotation invariance. See the Appendix for the results on (rotated) 3DMatch.}
\label{tab:ablation}
\vspace{-0.55cm}
\end{table}

\noindent \textbf{The Number of Global Transformers.} To demonstrate the importance of being globally aware, we first remove the global transformer. The substantial performance drop confirms the significance of global awareness. Then we add one global transformer and observe an increased performance. In our default setting with 3 global transformers, the model performs the best. However, when the number is increased to 5, the model shows a slight performance drop, which we owe to overfitting. As the data augmentation of rotations has less effect on an intrinsically rotation-invariant method, more data is required for training a larger model.

\vspace{-0.2cm}
\section{Conclusion}
\label{sec:conclusion}
We introduced \OURS{} - an intrinsically rotation-invariant model for point cloud matching. We proposed PAM~(PPF Attention Mechanism) that embeds PPF-based local coordinates to encode rotation-invariant geometry. This design lies at the core of AAL~(Attention Abstraction Layer), PAL~(PPF Attention Layer), and TUL~(Transition Up Layer) which are consecutively stacked to compose PPFTrans~(PPF Transformer) for representative and pose-agnostic geometry description. We further enhanced features by introducing a novel global transformer architecture, which ensures the rotation-invariant cross-frame spatial awareness.
Extensive experiments are conducted on both rigid and non-rigid benchmarks to demonstrate the superiority of our approach, especially the remarkable robustness against arbitrary rotations. 
Limitations are discussed in the Appendix.

\noindent\textbf{Acknowledgment.} This paper is supported by the National Natural Science Foundation of China under Grant No. 62025208. We appreciate the help from Lennart Bastian, Mert Karaoglu, Ning Liu, and Zhiying Leng.

\section{Appendix}
In this appendix, we first provide the implementation details in Sec.~\ref{sec:implementation}. Then the network architecture is detailed in Sec.~\ref{sec:network}. Moreover, the detailed definition of the geometric embedding, the loss function, and the evaluation metrics are illustrated in Sec.~\ref{sec:embedding}, Sec.~\ref{sec:loss}, and Sec.~\ref{sec:metrics}, respectively. Furthermore, more quantitative and qualitative results are demonstrated in Sec.~\ref{sec:more_quantitative} and Sec.~\ref{sec:more_qualitative}, respectively. Finally, the runtime is first analyzed in Sec.~\ref{sec:runtime}, and the limitations are further discussed in Sec.~\ref{sec:limitations}.

\subsection{Implementation Details}
\label{sec:implementation}
We implement \OURS{} with PyTorch~\cite{paszke2019pytorch}. The matching model can be trained end-to-end on a single Nvidia RTX 3090 with 24G memory. In practice, we train the model on-parallel using 4$\times$ Nvidia 3090 GPUs for $\sim$35 epochs on both 3DMatch/3DLoMatch~\cite{zeng20173dmatch,huang2021predator} and 4DMatch/4DLoMatch~\cite{li2022lepard}. It takes $\sim$35 hours and $\sim$30 hours for full convergence on 3DMatch/3DLoMatch and 4DMatch/4DLoMatch, respectively. The batch size is set to 1. We use an Adam optimizer~\cite{kingma2014adam} with an initial learning rate of 1e-4, which is exponentially decayed by 0.05 after each epoch. On 3DMatch/3DLoMatch, we select $|\mathcal{C}^\prime|=256$ superpoint correspondences with the highest scores. Based on each superpoint correspondence, we further extract the mutual top-3 point correspondences whose confidence scores are larger than 0.05 as the point correspondences. For non-rigid matching, we first pick the superpoint correspondences whose Euclidean distance is smaller than 0.75~(pick the top-128 instead if the number of selected correspondences is smaller than 128) and extract the mutual top-2 point correspondences with scores larger than 0.05. 

\renewcommand\arraystretch{1}
\begin{table*}[ht!]
\centering
\resizebox{0.7\textwidth}{!}{
\begin{tabular}{c|c|c}
\toprule
Stage &Block &Operation  \\
\midrule
\midrule
Input & &$\mathcal{P}=(\mathbf{P}, \mathbf{N}, \mathbf{X}\in \mathbb{R}^{n\times1})$  \\
\midrule
\multirow{8}{*}{Encoder} &\multirow{2}{*}{$\text{Block}^e_{1}(\mathcal{P})\rightarrow\mathcal{P}_1$} &AAL$(n \times 1) \rightarrow n\times 64$\\

& &PAL$(n \times 64) \rightarrow n\times 64$\\
\cline{2-3}

&\multirow{2}{*}{$\text{Block}^e_{2}(\mathcal{P}_1)\rightarrow\mathcal{P}_2$} &AAL$(n \times 64) \rightarrow n / 4\times 128$ \\

& &PAL$(n / 4 \times 128)\rightarrow n / 4\times 128$\\
\cline{2-3}

&\multirow{2}{*}{$\text{Block}^e_{3}(\mathcal{P}_2)\rightarrow\mathcal{P}_3$} &AAL$(n / 4 \times 128) \rightarrow n / 16\times 256$ \\

& &PAL$(n / 16 \times 256) \rightarrow n / 16\times 256$\\
\cline{2-3}

&\multirow{2}{*}{$\text{Block}^e_{4}(\mathcal{P}_3)\rightarrow\mathcal{P}^\prime$} &AAL$(n / 16 \times 256) \rightarrow n / 64\times 256$  \\

& &PAL$(n / 64 \times 256) \rightarrow n / 64\times 256$ \\
\midrule
\multirow{8}{*}{Decoder} &\multirow{2}{*}{$\text{Block}^d_{4}(\mathcal{P}^\prime)\rightarrow\hat{\mathcal{P}}_4$} &TUL $(n/64 \times 256) \rightarrow n/64\times 256$\\

& &PAL: $n/64\times 256 \rightarrow n/64\times 256$\\
\cline{2-3}

&\multirow{2}{*}{$\text{Block}^d_{3}(\hat{\mathcal{P}}_4, \mathcal{P}_3)\rightarrow\hat{\mathcal{P}}_3$} &TUL$(n/64 \times 256,\; n/16 \times 256)\rightarrow n / 16\times 256$  \\

& &PAL$(n / 16 \times 256)\rightarrow n / 16\times 256$ \\
\cline{2-3}

&\multirow{2}{*}{$\text{Block}^d_{2}(\hat{\mathcal{P}}_3, \mathcal{P}_2)\rightarrow\hat{\mathcal{P}}_2$} &TUL$(n / 16 \times 256,\;n / 4\times 128)\rightarrow n / 4\times 128$  \\

& &PAL$(n / 4 \times 128) \rightarrow n / 4\times 128$ \\
\cline{2-3}

&\multirow{2}{*}{$\text{Block}^d_{1}(\hat{\mathcal{P}}_2, \mathcal{P}_1)\rightarrow\hat{\mathcal{P}}$} &TUL$(n / 4 \times 128,\; n\times 64)\rightarrow n\times 64$ \\

& &PAL$(n \times 64)\rightarrow n \times 64$\\
\midrule
Output & &$\mathcal{P}^\prime=(\mathbf{P}^\prime, \mathbf{N}^\prime, \mathbf{X}^\prime); \quad \hat{\mathcal{P}}=(\hat{\mathbf{P}}, \hat{\mathbf{N}},\hat{\mathbf{X}})$  \\
\bottomrule

\end{tabular}
}
\caption{Detailed architecture of the PPFTrans encoder-decoder.}
\label{tab:arch_local}
\end{table*}

\subsection{Network Architecture}
\label{sec:network}
\paragraph{PPFTrans Encoder-Decoder.} We detail the architecture of PPFTrans in Tab.~\ref{tab:arch_local}. The encoder part has $4\times$ encoder blocks. In each block, AAL first downsamples the points and aggregates the information in a local vicinity. PAL further enhances the features with both the pose-agnostic local geometry and highly-representative learned context. The decoder part also comprises $4\times$ decoder blocks. In each block~(except for $\text{Block}_4$), TUL first subsamples the points and incorporates the information flowing from the encoder via skip connections. The obtained features are further enhanced by the following PAL.

\paragraph{Global Transformer.} The details of the global transformer are demonstrated in Tab.~\ref{tab:arch_global}. It has $3\times$ transformer blocks, each comprising a geometry-aware self-attention module~(GSM) followed by a position-aware cross-attention module~(PCM). In each transformer block, GSM first aggregates the global context individually for each point cloud. Then in PCM, the global context flows from the second frame to the first one and then from the first frame to the second one.

\begin{figure}
  \centering
  \includegraphics[width=0.38\textwidth]{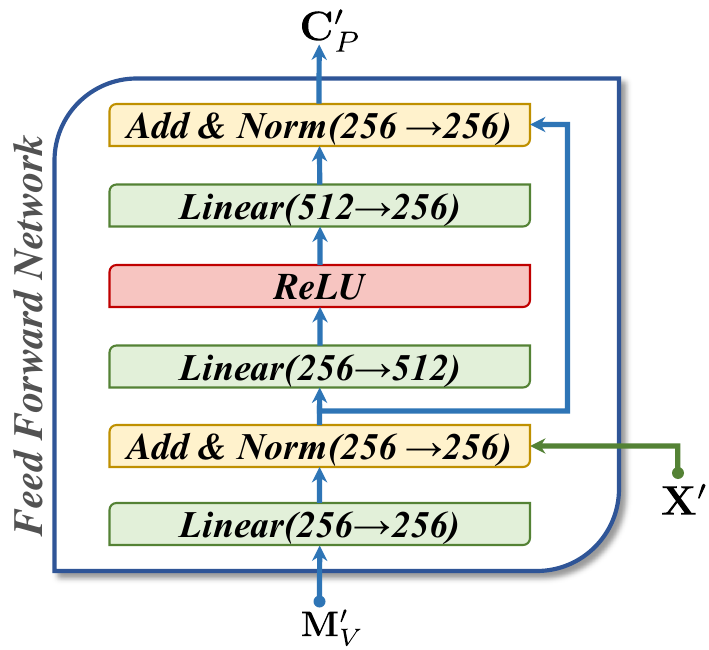}
  \caption{Detailed architecture of the feed-forward network. LayerNorm~\cite{ba2016layer} is used for normalization.}
  \label{fig:feedforward}
\end{figure}

\noindent\textbf{Feed-Forward Network.} The structure of the feed-forward network is illustrated in Fig.~\ref{fig:feedforward}. It details the feed-forward network in the context branch of GSM in Fig.~4 of the main paper.

\renewcommand\arraystretch{1}
\begin{table*}[ht!]
\centering
\resizebox{0.75\textwidth}{!}{
\begin{tabular}{c|c|c|c|c}
\toprule
Block &\multicolumn{2}{c|}{Module} &\multicolumn{2}{c}{Operation}  \\
\midrule
\midrule
Input & & &$\mathcal{P}^\prime=(\mathbf{P}^\prime, \mathbf{N}^\prime, \mathbf{X}^\prime)$ &$\mathcal{Q}^\prime=(\mathbf{Q}^\prime, \mathbf{M}^\prime, \mathbf{Y}^\prime)$  \\
\midrule
\multirow{3}{*}{$\text{Trans}_1$} &$\text{Self}_1(\mathcal{P}^\prime)\rightarrow\widetilde{\mathcal{P}}^\prime_1$
&$\text{Self}_1(\mathcal{Q}^\prime)\rightarrow\widetilde{\mathcal{Q}}^\prime_1$ 
&GSM$(n^\prime\times c^\prime) \rightarrow n^\prime\times c^\prime$ &GSM$(m^\prime\times c^\prime) \rightarrow m^\prime\times c^\prime$\\
&\multicolumn{2}{c|}{$\text{Cross}_1(\widetilde{\mathcal{P}}^\prime_1, \widetilde{\mathcal{Q}}^\prime_1)\rightarrow\mathcal{P}^\prime_1$} &\multicolumn{2}{c}{PCM$(n^\prime\times c^\prime,m^\prime\times c^\prime) \rightarrow n^\prime\times c^\prime$}\\
&\multicolumn{2}{c|}{$\text{Cross}_1(\widetilde{\mathcal{Q}}^\prime_1, {\mathcal{P}}^\prime_1)\rightarrow\mathcal{Q}^\prime_1$} &\multicolumn{2}{c}{PCM$(m^\prime\times c^\prime,n^\prime\times c^\prime) \rightarrow m^\prime\times c^\prime$}\\
\midrule
\multirow{3}{*}{$\text{Trans}_2$} &$\text{Self}_2(\mathcal{P}^\prime_1)\rightarrow\widetilde{\mathcal{P}}^\prime_2$
&$\text{Self}_2(\mathcal{Q}^\prime_1)\rightarrow\widetilde{\mathcal{Q}}^\prime_2$ 
&GSM$(n^\prime\times c^\prime) \rightarrow n^\prime\times c^\prime$ &GSM$(m^\prime\times c^\prime) \rightarrow m^\prime\times c^\prime$\\
&\multicolumn{2}{c|}{$\text{Cross}_2(\widetilde{\mathcal{P}}^\prime_2, \widetilde{\mathcal{Q}}^\prime_2)\rightarrow\mathcal{P}^\prime_2$} &\multicolumn{2}{c}{PCM$(n^\prime\times c^\prime,m^\prime\times c^\prime) \rightarrow n^\prime\times c^\prime$}\\
&\multicolumn{2}{c|}{$\text{Cross}_2(\widetilde{\mathcal{Q}}^\prime_2, {\mathcal{P}}^\prime_2)\rightarrow\mathcal{Q}^\prime_2$} &\multicolumn{2}{c}{PCM$(m^\prime\times c^\prime,n^\prime\times c^\prime) \rightarrow m^\prime\times c^\prime$}\\
\midrule
\multirow{3}{*}{$\text{Trans}_3$} &$\text{Self}_3(\mathcal{P}^\prime_2)\rightarrow\widetilde{\mathcal{P}}^\prime_3$
&$\text{Self}_3(\mathcal{Q}^\prime_2)\rightarrow\widetilde{\mathcal{Q}}^\prime_3$ 
&GSM$(n^\prime\times c^\prime) \rightarrow n^\prime\times c^\prime$ &GSM$(m^\prime\times c^\prime) \rightarrow m^\prime\times c^\prime$\\
&\multicolumn{2}{c|}{$\text{Cross}_3(\widetilde{\mathcal{P}}^\prime_3, \widetilde{\mathcal{Q}}^\prime_3)\rightarrow\widetilde{\mathcal{P}}^\prime$} &\multicolumn{2}{c}{PCM$(n^\prime\times c^\prime,m^\prime\times c^\prime) \rightarrow n^\prime\times c^\prime$}\\
&\multicolumn{2}{c|}{$\text{Cross}_3(\widetilde{\mathcal{Q}}^\prime_3, {\mathcal{P}}^\prime_3)\rightarrow\widetilde{\mathcal{Q}}^\prime$} &\multicolumn{2}{c}{PCM$(m^\prime\times c^\prime,n^\prime\times c^\prime) \rightarrow m^\prime\times c^\prime$}\\
\midrule
Output & & &$\widetilde{\mathcal{P}}^\prime=(\mathbf{P}^\prime, \mathbf{N}^\prime, \widetilde{\mathbf{X}}^\prime)$ &$\widetilde{\mathcal{Q}}^\prime=(\mathbf{Q}^\prime, \mathbf{M}^\prime, \widetilde{\mathbf{Y}}^\prime)$  \\
\bottomrule

\end{tabular}
}
\caption{Detailed architecture of the global transformer.}
\label{tab:arch_global}
\end{table*}

\subsection{Geometric Embedding}
\label{sec:embedding}
Taking superpoints $\mathbf{P}^\prime \in \mathbb{R}^{n^\prime \times 3}$ as an instance, the geometric embedding $\mathbf{G}^\prime_P\in \mathbb{R}^{n^\prime \times n^\prime \times c^\prime}$ proposed in~\cite{qin2022geometric} depicts the pairwise geometric relationship among superpoints in a rotation-invariant fashion. It comprises a distance-based part $\mathbf{G}^\prime_D\in \mathbb{R}^{n^\prime\times n^\prime \times c^\prime}$ as well as an angle-based part $\mathbf{G}^\prime_A\in \mathbb{R}^{n^\prime\times n^\prime \times 3\times c^\prime}$.

\paragraph{Euclidean Distance.} The pairwise Euclidean distance is defined as $\rho_{i, j} = \lVert\mathbf{p}^\prime_i - \mathbf{p}^\prime_j\rVert_2$, which is projected to a $c^\prime$-dimension~(note that $c^\prime$ must be an even number) embedding via the sinusoidal function~\cite{vaswani2017attention}:
\begin{equation}
\left\{
\begin{array}{r}
      \mathbf{G}^\prime_D(i, j, 2l + 1) = \sin(\frac{\rho_{i, j}/\sigma_d}{10000^{2l/c^\prime}}),\\
       \mathbf{G}^\prime_D(i, j, 2l + 2) = \cos(\frac{\rho_{i, j}/\sigma_d}{10000^{2l/c^\prime}}),
\end{array}
\right.
\label{eq:geo_distance}
\end{equation}
with $0\leq l < c^\prime/2$ and $\sigma_d=0.2$.

\paragraph{Angles.} Given a superpoint pair $(\mathbf{p}^\prime_i, \mathbf{p}^\prime_j)$, the $3$-nearest neighbors of $\mathbf{p}^\prime_i$ w.r.t. $\mathbf{P}^\prime$ is first retrieved and denoted as $\mathcal{N}(i)$. For each $k\in \mathcal{N}(i)$, we calculate the angle between two vectors by $\alpha^k_{i, j} = \angle(\mathbf{p^\prime_k} - \mathbf{p^\prime_i}, \mathbf{p^\prime_j} - \mathbf{p^\prime_i})$~\cite{birdal2015point,deng2018ppfnet}, upon which the $c^\prime$-dimension angle-based embedding is defined as:
\begin{equation}
\left\{
\begin{array}{r}
      \mathbf{G}^\prime_A(i, j, k, 2l + 1) = \sin(\frac{\alpha^k_{i, j}/\sigma_a}{10000^{2l/c^\prime}}),\\
       \mathbf{G}^\prime_A(i, j, k, 2l + 2) = \cos(\frac{\alpha^k_{i, j}/\sigma_a}{10000^{2l/c^\prime}}),
\end{array}
\right.
\label{eq:geo_angle}
\end{equation}
with $0\leq l < c^\prime/2$ and $\sigma_a=15$.

The pairwise geometric embedding $\mathbf{G}^\prime_P$ finally reads as:

\begin{equation}
\mathbf{G}^\prime_P = \mathbf{G}^\prime_D\mathbf{W}_D + \max\limits_{k}(\mathbf{G}^\prime_A\mathbf{W}_A),
\label{eq:geo}
\end{equation}
where $\max\limits_{k}(\mathbf{G}^\prime_A\mathbf{W}_A)$ indicates the max-pooling operation over the second last dimension, and $\mathbf{W}_D, \mathbf{W}_A\in \mathbb{R}^{c^\prime\times c^\prime}$ stand for two learnable matrices.

\subsection{Loss Function}
\label{sec:loss}
\paragraph{Superpoint Matching Loss.} We use the Circle Loss~\cite{sun2020circle} for superpoint matching. Following~\cite{qin2022geometric}, we use the overlap ratio between the vicinity of superpoints to weigh different ground truth superpoint correspondences. More specifically, for each superpoint $\mathbf{p}^\prime_i\in \mathbf{P}^\prime$ with an associated feature $\widetilde{\mathbf{x}}^\prime_i$~(unit vector after normalization), we sample a positive set of superpoints from $\mathbf{Q}^\prime$, denoted as $\mathcal{E}^P_i = \{\mathbf{q}^\prime_j \in \mathbf{Q}^\prime | \mathcal{O}(\mathbf{p}^\prime_i, \mathbf{q}^\prime_j) > \tau_r\}$, where $\mathcal{O}$ is the function that calculates the overlap ratio between the vicinity of two superpoints, and $\tau_r$ is the threshold to select positive samples~($\tau_r$=0.1 by default). The overlap ratio function is defined as:

\begin{equation}
\mathcal{O}(\mathbf{p}^\prime_i, \mathbf{q}^\prime_j) = \frac{|\{\hat{\mathbf{p}}_u\in\hat{\mathbf{G}}^P_i|\exists\hat{\mathbf{q}}_v\in \hat{\mathbf{G}}^Q_j \;\text{s.t.}\; \hat{\mathbf{p}}_u \Leftrightarrow \hat{\mathbf{q}}_v\}|}{|\{\hat{\mathbf{p}}_u \in \hat{\mathbf{G}}^P_i\}|},
\label{eq:overlap_ratio}
\end{equation}
\noindent where $\Leftrightarrow$ denotes the correspondence relationship and $\hat{\mathbf{G}}^P_i$ is the group of points from $\hat{\mathbf{P}}$ assigned to $\mathbf{p}^\prime_i$ by the Point-to-Node grouping strategy~\cite{yu2021cofinet}.

We further sample a negative set of superpoints $\mathcal{F}^P_i = \{\mathbf{q}^\prime_j \in \mathbf{Q}^\prime | \mathcal{O}(\mathbf{p}^\prime_i, \mathbf{q}^\prime_j) =0\}$. Then for $\mathbf{P}^\prime$, the superpoint matching loss is computed as:

\begin{equation}
\begin{split}
    \mathcal{L}^P_c = \frac{1}{n^\prime}&\sum_{i=1}^{n^\prime}\text{log}[1 + \\
    &\sum_{\mathbf{q}^\prime_j \in \mathcal{E}^P_i}{e^{r^j_i\beta^{i, j}_e(d^j_i-\Delta_e)}}\cdot \sum_{\mathbf{q}^\prime_k\in \mathcal{F}^P_i}{e^{\beta^{i, k}_f(\Delta_f-d^k_i)}}],
\end{split}
\label{eq:coarse}
\end{equation}

\noindent with $r^j_i := \mathcal{O}(\mathbf{p}^\prime_i, \mathbf{q}^\prime_j)$ and $d^j_i = \lVert\widetilde{\mathbf{x}}^\prime_i - \widetilde{\mathbf{y}}^\prime_j \rVert_2$. Moreover, $\Delta_e$ and $\Delta_f$ are the positive and negative margins~($\Delta_e$=0.1 and $\Delta_f$=1.4 by default). $\beta^{i, j}_e = \gamma(d^j_i - \Delta_e)$ and $\beta^{i, k}_f=\gamma(\Delta_f - d^k_i)$ are the weights individually determined for different samples, with $\gamma$ a hyper-parameter. The same loss for $\mathbf{Q}^\prime$ is defined in a similar way, and the overall loss reads as $\mathcal{L}_s = (\mathcal{L}^P_s + \mathcal{L}^Q_s)/2$.

\paragraph{Point Matching Loss.} For each superpoint correspondence $\mathcal{C}^\prime_l = (\mathbf{p}^\prime_i, \mathbf{q}^\prime_j) \in \mathcal{C}^\prime$, we adopt a negative log-likelihood loss~\cite{sarlin2020superglue} on its corresponding normalized similarity matrix $\overline{\mathbf{C}}_l\in \mathbb{R}^{(\overline{n} + 1)\times (\overline{m} + 1)}$. We define $\mathcal{M}_l = \{(u, v)| \hat{\mathbf{p}}_u \Leftrightarrow \hat{\mathbf{q}}_v \;\text{with}\;\hat{\mathbf{p}}_u \in \hat{\mathbf{G}}^P_i, \hat{\mathbf{q}}_v \in \hat{\mathbf{G}}^Q_j\}$ as the set comprising the indices of corresponding points. We further define $\mathcal{I}_l = \{u|\hat{\mathbf{p}}_u \nLeftrightarrow \hat{\mathbf{q}}_v, \forall\hat{\mathbf{q}}_v \in \hat{\mathbf{G}}^Q_j\}$ and $\mathcal{J}_l = \{v|\hat{\mathbf{q}}_v \nLeftrightarrow \hat{\mathbf{p}}_u, \forall\hat{\mathbf{p}}_u \in \hat{\mathbf{G}}^P_i\}$ as the sets of indices of points that have no correspondence in the opposite frame, with $\nLeftrightarrow$ depicting non-correspondence relationship. Then the point matching loss on $\mathcal{C}^\prime_l$ can be defined as:

\begin{equation}
\begin{split}
\mathcal{L}_f^l = -\sum_{(u, v)\in \mathcal{M}_l} \log \overline{c}^l_{u, v} &- \sum_{u\in \mathcal{I}_l} \log \overline{c}^l_{u, \overline{m} + 1} \\ &- \sum_{v \in \mathcal{J}_l} \log \overline{c}^l_{\overline{n} + 1, v} \quad ,
\end{split}
\label{eq:fine}
\end{equation}
where $\overline{c}^{l}_{u, v}:= \overline{\mathbf{C}}_l(u, v)$ stands for the entry on the $u^{th}$ row  and $v^{th}$ column of $\overline{\mathbf{C}}_l$. The overall point matching loss reads as $\mathcal{L}_p = \frac{1}{|\mathcal{C}^\prime|}\sum_{l=1}^{|\mathcal{C}^\prime|} \mathcal{L}^l_p$.

\subsection{Detailed Metrics}
\label{sec:metrics}
Given a point cloud pair $\mathbf{P}\in \mathbb{R}^{n\times3}$ and $\mathbf{Q}\in \mathbb{R}^{m \times 3}$, \OURS{} generates a correspondence set $\mathcal{C}$ by matching the downsampled point cloud pair $\hat{\mathbf{P}}\in \mathbb{R}^{\hat{n}\times 3}$ and $\hat{\mathbf{Q}}\in \mathbb{R}^{\hat{m}\times 3}$. We detail all the metrics for evaluation hereafter.

\paragraph{Inlier Ratio~(IR).} IR counts the fraction of putative correspondences $(\hat{\mathbf{p}}_u, \hat{\mathbf{q}}_v)\in \mathcal{C}$ whose Euclidean distance is under a threshold $\tau_1$~(0.1m on 3DMatch/3DLoMatch, 0.04m on 4DMatch/4DLoMatch) under the ground-truth transformation $\mathbf{T}^*$:

\begin{equation}
\mathcal{I}(\mathcal{C}\big| \mathbf{T}^*) = \frac{1}{|\mathcal{C}|}\sum_{(\hat{\mathbf{p}}_u, \hat{\mathbf{q}}_v)\in \mathcal{C}} \mathds{1}(\lVert\mathbf{T}^*(\hat{\mathbf{p}}_u) - \hat{\mathbf{q}}_v\rVert_2 < \tau_1),
\label{eq:ir}
\end{equation}
with $\mathds{1}(\cdot)$ the indicator function.

\paragraph{Feature Matching Recall~(FMR).} FMR counts the fraction of point cloud pairs whose IR is larger than a threshold $\tau_2=0.05$:
\begin{equation}
\mathcal{F}(\mathcal{T}) = \frac{1}{|\mathcal{T}|}\sum_{t=1}^{|\mathcal{T}|}\mathds{1}(\mathcal{I}(\mathcal{C}_t \big| \mathbf{T}^*_t) > \tau_2),
\label{eq:fmr}
\end{equation}
with $\mathcal{T}$ the testing set and $\mathcal{T}_t$ the $t^{th}$ point cloud pair in the dataset.

\paragraph{Registration Recall~(RR).} RR computes the fraction of point cloud pairs that are registered correctly based on the putative correspondences, measured by the \textit{Root-Mean-Square Error}~(RMSE). Following~\cite{huang2021predator}, we define RMSE on the original 3DMatch/3DLoMatch as:

\begin{equation}
\mathcal{R}_1(\mathcal{C} \big| \mathcal{C}^*) = \sqrt{\frac{1}{|\mathcal{C}^*|}\sum_{(\mathbf{p}_i, \mathbf{q}_j)\in \mathcal{C}^*}\lVert\mathbf{T}(\mathbf{p}_i) - \mathbf{q}_j\rVert_2^2},
\label{eq:rmse1}
\end{equation}
with $\mathcal{C}^*$ the ground-truth correspondence set established upon $\mathbf{P}$ and $\mathbf{Q}$, and $\mathbf{T}$ the transformation estimated based on $\mathcal{C}$.
On Rotated 3DMatch/3DLoMatch, we follow~\cite{yuan2020deepgmr,yu2022riga} and define the RMSE as:

\begin{equation}
\mathcal{R}_2(\mathcal{C}\big|\mathbf{T}^*, \mathbf{P}) \approx \frac{1}{n}\sqrt{\sum_{\mathbf{p}_i\in \mathbf{P}}\lVert\mathbf{T}(\mathbf{p}_i) - \mathbf{T}^*(\mathbf{p}_i)\rVert_2^2},
\label{eq:rmse2}
\end{equation}
with $\mathbf{T}$ the transformation estimated based on $\mathcal{C}$ and $\mathbf{T}^*$ the ground-truth transformation. RR is finally calculated as:
\begin{equation}
\begin{split}
\mathcal{R}(\mathcal{T}) = &\frac{1}{|\mathcal{T}|}\sum_{t=1}^{\mathcal{T}}\mathds{1}(\mathcal{R}_1(\mathcal{C} \big| \mathcal{C}^*) < \tau_3) \quad \text{or}\\
                            &\frac{1}{|\mathcal{T}|}\sum_{t=1}^{\mathcal{T}}\mathds{1}(\mathcal{R}_2(\mathcal{C}\big|\mathbf{T}^*, \mathbf{P})) < \tau_3),\\
\end{split}
\label{eq:rr}
\end{equation}
\noindent with $\tau_3$ = 0.2m.

\paragraph{Non-Rigid Feature Matching Recall~(NFMR).} NFMR counts the fraction of ground-truth correspondences $\mathcal{C}^*$ that can be recovered by the putative correspondences $\mathcal{C}$. The deformation flow $\hat{\mathbf{d}}_u$ for each putative correspondence $(\hat{\mathbf{p}}_u, \hat{\mathbf{q}}_v)\in \mathcal{C}$ is defined as $\hat{\mathbf{d}}_u = \hat{\mathbf{q}}_v - \hat{\mathbf{p}}_u$. Then for each $(\mathbf{p}_i, \mathbf{q}_j)\in \mathcal{C}^*$, the deformation flow can be computed via interpolation:

\begin{equation}
\mathbf{d}_i = \frac{\sum_{u \in \mathcal{N}(i)}w_u^i\hat{\mathbf{d}}_u}{\sum_{u\in \mathcal{N}(i)}w_u^i}, \; \text{with} \; w_u^i=\frac{1}{\lVert\mathbf{p}_i - \hat{\mathbf{p}}_u\rVert_2},
\label{eq:flow}
\end{equation}
\noindent where $\mathcal{N}(i)$ indicates the $k$-nearest neighbor~($k=3$ in practice) of $\mathbf{p}_i$ w.r.t. points $\hat{\mathbf{p}}_u$ s.t. $(\hat{\mathbf{p}}_u, \hat{\mathbf{q}}_v)\in \mathcal{C}$. NFMR is finally computed by:
\begin{equation}
\mathcal{F}_N(\mathcal{C}^*\big| \mathcal{C}) = \frac{1}{|\mathcal{C}^*|}\sum_{(\mathbf{p}_i, \mathbf{q}_j)\in \mathcal{C}^*}\mathds{1}(\lVert\mathbf{d}_i - \mathbf{d}_i^*\rVert_2 < \tau_4),
\label{eq:nfmr}
\end{equation}
\noindent with $\mathbf{d}_i^*$ the ground-truth deformation flow and $\tau_4$=0.04m in practice.

\subsection{More Quantitative Results}
\label{sec:more_quantitative}
\noindent\textbf{Varying Correspondence Number on Rotated Data.} We further analyze the performance of different methods w.r.t. the varying number of correspondences on rotated data. The superiority of \OURS{} can be observed in Tab.~\ref{tab:scene_rotated}. 

\renewcommand\arraystretch{0.95}
\begin{table}[ht!]
\centering

\resizebox{0.48\textwidth}{!}{
\begin{tabular}{lccccc|ccccc}
\toprule
 &\multicolumn{5}{c}{\textbf{Rotated 3DMatch}}  &\multicolumn{5}{c}{\textbf{Rotated 3DLoMatch}}\\
\# Samples &5000 &2500 &1000 &500 &250 &5000 &2500 &1000 &500 &250 \\
\midrule
\midrule
&\multicolumn{10}{c}{\textit{Feature Matching Recall} (\%) $\uparrow$}\\
\midrule
SpinNet~\cite{ao2021spinnet} &97.4 &97.4 &96.7 &96.5 &94.1 &75.2 &74.9 &72.6 &69.2 &61.8\\
Predator~\cite{huang2021predator} &96.2 &96.2 &96.6 &96.0 &96.0 &73.7 &74.2 &75.0 &74.8 &73.5\\
CoFiNet~\cite{yu2021cofinet} &97.4 &97.4 &97.2 &97.2 &97.3 &78.6 &78.8 &79.2 &78.9 &79.2\\

YOHO~\cite{wang2022you} &97.8 &97.8 &97.4 &97.6 &96.4 &77.8 &77.8 &76.3 &73.9 &67.3\\
RIGA~\cite{yu2022riga} &\textbf{98.2} &\textbf{98.2} &\textbf{98.2} &\underline{98.0} &\textbf{98.1} &84.5 &84.6 &84.5 &84.2 &84.4\\
GeoTrans~\cite{qin2022geometric} &97.8 &97.9 &98.1 &97.7 &97.3 &\underline{85.8} &\underline{85.7} &\underline{86.5} &\underline{86.6} &\underline{86.1}\\
\OURS{}~(\textit{Ours}) &\textbf{98.2} &\underline{98.1} &\underline{98.1} &\textbf{98.1} &\textbf{98.1} &\textbf{89.4} &\textbf{89.2} &\textbf{89.1} &\textbf{89.1} &\textbf{89.0}\\
\midrule
&\multicolumn{10}{c}{\textit{Inlier Ratio} (\%) $\uparrow$}\\
\midrule
SpinNet~\cite{ao2021spinnet} &48.7 &46.0 &40.6 &35.1 &29.0 &25.7 &23.9 &20.8 &17.9 &15.6\\
Predator~\cite{huang2021predator} &52.8 &53.4 &52.5 &50.0 &45.6  &22.4 &23.5 &23.0 &23.2 &21.6\\
CoFiNet~\cite{yu2021cofinet} &46.8 &48.2 &49.0 &49.3 &49.3 &21.5 &22.8 &23.6 &23.8 &23.8\\
YOHO~\cite{wang2022you} &64.1 &60.4 &53.5 &46.3 &36.9 &23.2 &23.2 &19.2 &15.7 &12.1 \\
RIGA~\cite{yu2022riga}  &\underline{68.5} &69.8 &70.7 &71.0 &71.2 &32.1  &33.5 &34.3 &34.7 &35.0\\
GeoTrans~\cite{qin2022geometric} &68.2 &\underline{72.5} &\underline{73.3} &\underline{79.5} &\underline{82.3} &\underline{40.0} &\underline{40.3} &\underline{42.7} &\underline{49.5} &\underline{54.1}\\
\OURS{}~(\textit{Ours}) &\textbf{82.3} &\textbf{82.3} &\textbf{82.6} &\textbf{82.6} &\textbf{82.6} &\textbf{53.2} &\textbf{54.9} &\textbf{55.1} &\textbf{55.2} &\textbf{55.3}\\
\midrule

&\multicolumn{10}{c}{\textit{Registration Recall} (\%) $\uparrow$}\\
\midrule
SpinNet~\cite{ao2021spinnet} &93.2 &93.2 &91.1 &87.4 &77.0 &61.8 &59.1 &53.1 &44.1 &30.7\\
Predator~\cite{huang2021predator}&92.0 &92.8 &92.0 &92.2 &89.5 &58.6 &59.5 &60.4 &58.6 &55.8\\
CoFiNet~\cite{yu2021cofinet} &92.0 &91.4 &91.0 &90.3 &89.6 &62.5 &60.9 &60.9 &59.9 &56.5\\

YOHO~\cite{wang2022you}&92.5 &92.3 &92.4 &90.2 &87.4 &66.8 &67.1 &64.5 &58.2 &44.8\\

RIGA~\cite{yu2022riga} &\underline{93.0} &\underline{93.0} &\underline{92.6} &\underline{91.8} &\underline{92.3} &66.9 &67.6 &67.0 &66.5 &66.2\\
GeoTrans~\cite{qin2022geometric} &92.0 &91.9 &91.8 &91.5 &91.4 &\underline{71.8} &\underline{72.0} &\underline{72.0} &\underline{71.6} &\underline{70.9}\\
\OURS{}~(\textit{Ours}) &\textbf{94.7} &\textbf{94.9} &\textbf{94.4} &\textbf{94.4} &\textbf{94.2} &\textbf{77.2} &\textbf{76.5} &\textbf{76.6} &\textbf{76.5} &\textbf{76.0}\\
\bottomrule
\end{tabular}}
\caption{Quantitative results on Rotated 3DMatch \& 3DLoMatch with a varying number of points/correspondences.}
\label{tab:scene_rotated}

\end{table}

\paragraph{Ablation Study on (Rotated) 3DMatch.} We also conduct ablation study on (Rotated) 3DMatch, as shown in Tab.~\ref{tab:ablation_3dmatch}. Similar to the ablation study on (Rotated) 3DLoMatch shown in the main paper, our default model achieves the best performance, which further confirms the significance of each individual design of \OURS{}.

\renewcommand\arraystretch{0.9}
\begin{table}[ht!]
\centering
\resizebox{0.48\textwidth}{!}{
\begin{tabular}{ll|cccccc}
\toprule
& &\multicolumn{6}{c}{\textbf{3DMatch}}\\
\midrule
& &\multicolumn{3}{c|}{\textbf{Origin}} &\multicolumn{3}{c}{\textbf{Rotated}}\\
Category & Model &FMR &IR &RR  &FMR &IR &RR \\
\midrule
\midrule
\multirow{4}{*}{a. Local}
 &\;\ 1. PT~\cite{zhao2021point}&96.7 &71.0 &87.6 &96.4 &69.5 &90.5\\
&*2. PPF+PT~\cite{zhao2021point} &97.9 &80.1 &91.2 &97.8 &79.8 &93.9\\
& \; 3. $\Delta$xyz+Ours &-&-&-&-&-&-\\
& *4. \textit{Ours} &\textbf{98.0} &\textbf{82.6} &\textbf{91.9} &\textbf{98.2} &\textbf{82.3} &\textbf{94.7}\\
\midrule
\multirow{3}{*}{b. Aggregation}& *1. max pooling &97.9&80.8&90.7&97.8&80.8&94.1\\
& *2. avg pooling  &\textbf{98.1} &81.8&\textbf{92.1}&\textbf{98.2} &81.8 &\textbf{94.8}\\
& *3. \textit{Ours} &98.0 &\textbf{82.6} &91.9 &\textbf{98.2} &\textbf{82.3} &94.7\\
\midrule
\multirow{2}{*}{c. Backbone}&\;\ 1. KPConv~\cite{thomas2019kpconv} &97.9&74.6&91.1&97.3&72.8&94.3\\
& *2. \textit{Ours} &\textbf{98.0} &\textbf{82.6} &\textbf{91.9} &\textbf{98.2} &\textbf{82.3} &\textbf{94.7}\\
\midrule
\multirow{2}{*}{d. Global}&*1. GeoTrans~\cite{qin2022geometric} &97.9 &\textbf{82.6} &90.8 &98.0 &\textbf{82.3} &94.5\\
&*2. \textit{Ours} &\textbf{98.0} &\textbf{82.6} &\textbf{91.9} &\textbf{98.2} &\textbf{82.3} &\textbf{94.7}\\
\midrule
\multirow{4}{*}{e. \#Global}
& *1. $g=0$ &\textbf{98.5} &65.8 &90.9 &98.3 &65.9 &93.7  \\
& *2. $g=1$ &98.4 &74.8 &90.8 &\textbf{98.5} &74.8 &94.2 \\
& *3. $g=3$~(\textit{Ours}) &98.0 &\textbf{82.6} &\textbf{91.9} &98.2 &\textbf{82.3} &\textbf{94.7}\\
& *4. $g=5$ &98.1&82.0&91.7 &98.0 &82.0 &94.6\\

\bottomrule
\end{tabular}}
\caption{\small Ablation study on (Rotated) 3DMatch. 5,000 points/correspondences are leveraged. * indicates the methods with intrinsic rotation invariance.}
\label{tab:ablation_3dmatch}
\end{table}

\subsection{More Qualitative Results}
\label{sec:more_qualitative}

\begin{figure*}[h!]
  \centering
  \includegraphics[width=0.96\textwidth]{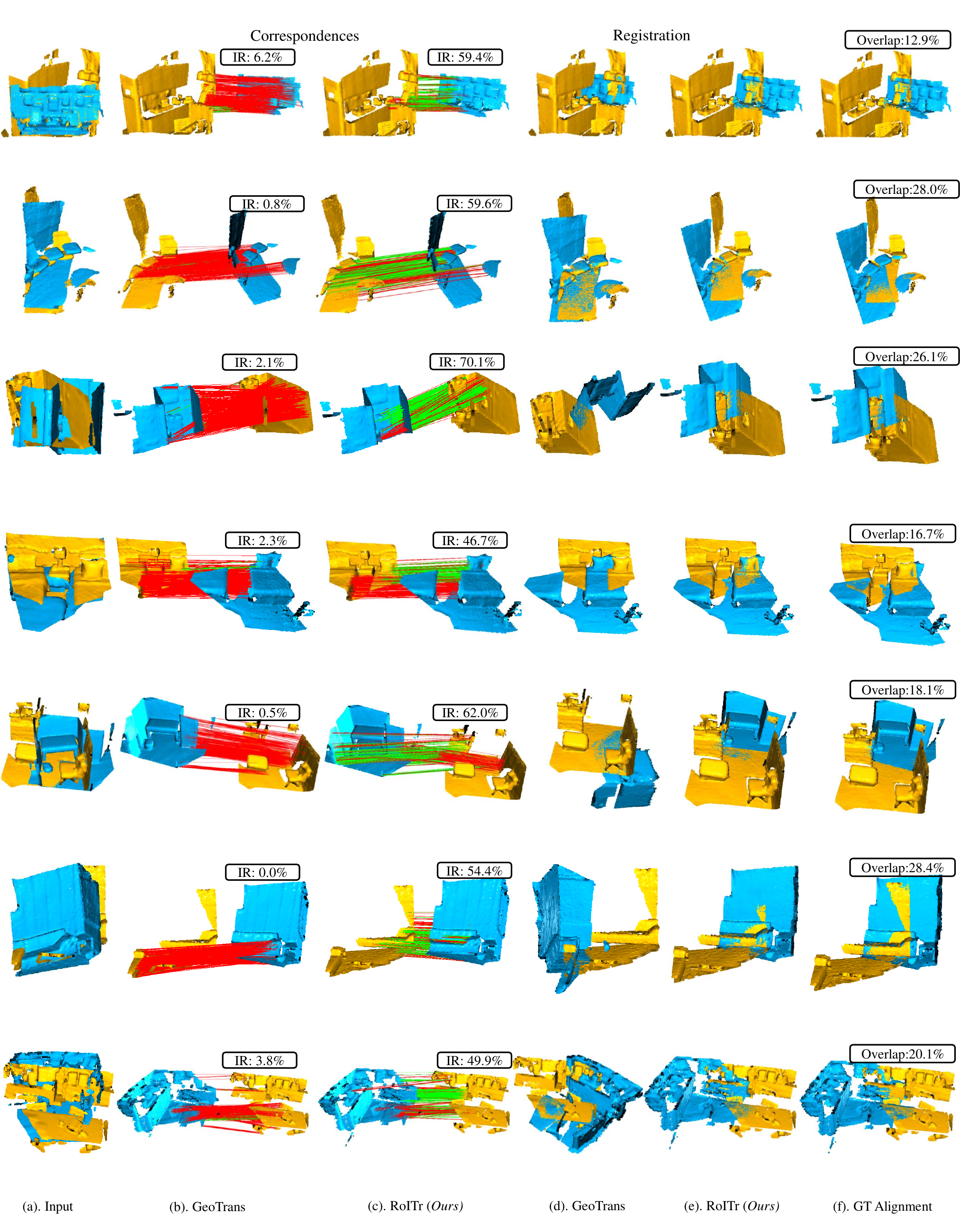}
  \caption{More qualitative results on 3DLoMatch. GeoTrans~\cite{qin2022geometric} is used as the baseline.}
  \label{fig:indoor_more}
\end{figure*}

\begin{figure*}[ht!]
  \centering
  \includegraphics[width=0.98\textwidth]{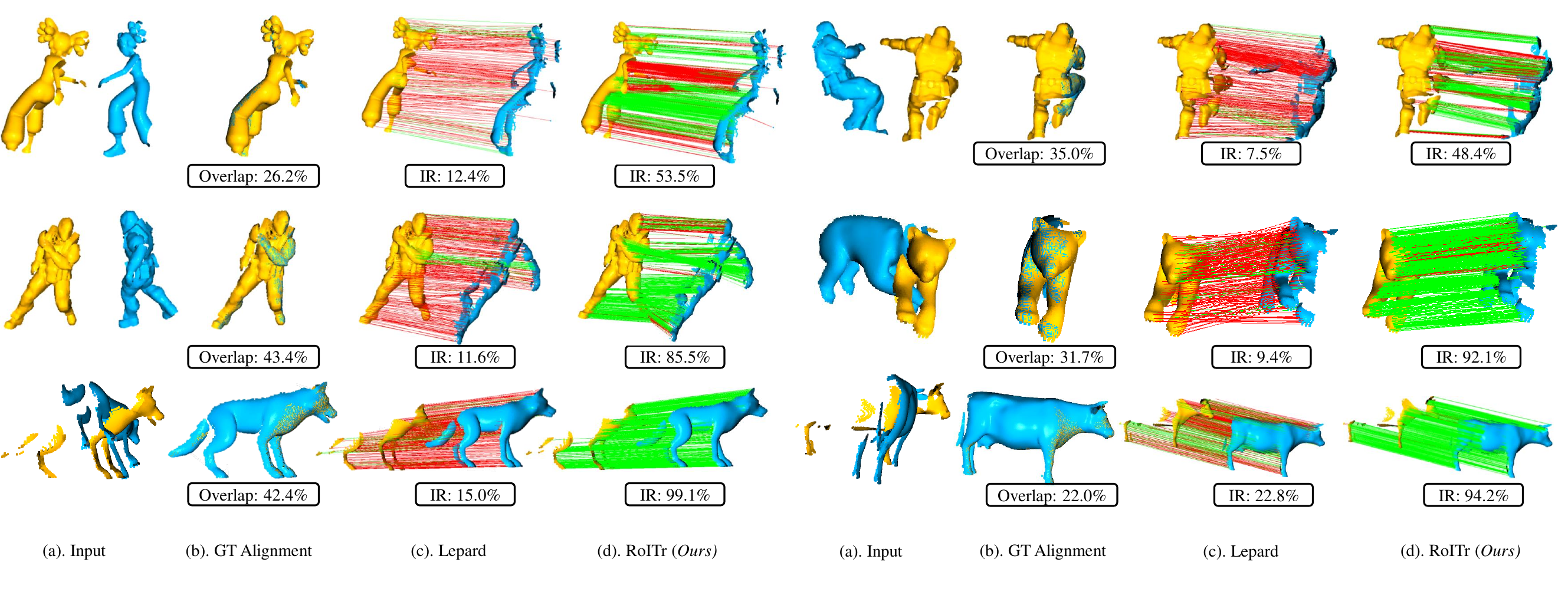}
  \caption{More qualitative results on 4DLoMatch. Lepard~\cite{li2022lepard} is used as the baseline.}
  \label{fig:deform_more}
\end{figure*}

\paragraph{Indoor Scenes: 3DLoMatch.} We show more qualitative results on the challenging 3DLoMatch benchmark in Fig.~\ref{fig:indoor_more}.
\noindent\textbf{Deformable Objects: 4DLoMatch.} More qualitative results of the 4DLoMatch benchmark consisting of partially-scanned deformable objects are demonstrated in Fig.~\ref{fig:deform_more}.
\begin{table}
\centering
\resizebox{0.39\textwidth}{!}{
\begin{tabular}{lccccc}
\toprule
Method &Data~(s)$\downarrow$ &Model~(s)$\downarrow$ &Total~(s)$\downarrow$\\
\hline
\hline
Lepard~\cite{li2022lepard} &0.444 &\textbf{0.051} &0.495\\
GeoTrans~\cite{qin2022geometric}  &\underline{0.194} &\underline{0.076} &\underline{0.270}\\
\OURS{}~(\textit{Ours}) &\textbf{0.023} &0.210 &\textbf{0.233}\\

\bottomrule
\end{tabular}
}
\caption{Runtime comparison.}
\label{tab:runtime}
\end{table}
\subsection{Runtime}
We show the runtime comparison with Lepard~\cite{li2022lepard} and GeoTrans~\cite{qin2022geometric} in Tab.~\ref{tab:runtime}. We run all the methods on a machine with a single Nvidia RTX 3090 GPU and an AMD Ryzen 5800X 3.80GHz CPU. All the models are tested without CPU parallel and with a batch size of 1. All the reported time is averaged over the 3DMatch testing set that consists of 1,623 point cloud pairs. The column "Data" counts the runtime for data preparation, and the column "Model" reports the time for generating descriptors from the prepared data. As shown in Tab.~\ref{tab:runtime}, \OURS{} has the highest data preparation and overall speed while the lowest model speed. That is mainly due to the relatively low speed of the attention mechanism compared to convolutions, e.g., KPConv~\cite{thomas2019kpconv} used in both Lepard and GeoTrans, and also because we do the Farthest Point Sampling~(FPS) and $k$-nearest neighbor search on GPU, which is counted into the model time.

\label{sec:runtime}

\subsection{Limitations}
\label{sec:limitations}
\begin{figure*}[h!]
  \centering
  \includegraphics[width=0.96\textwidth]{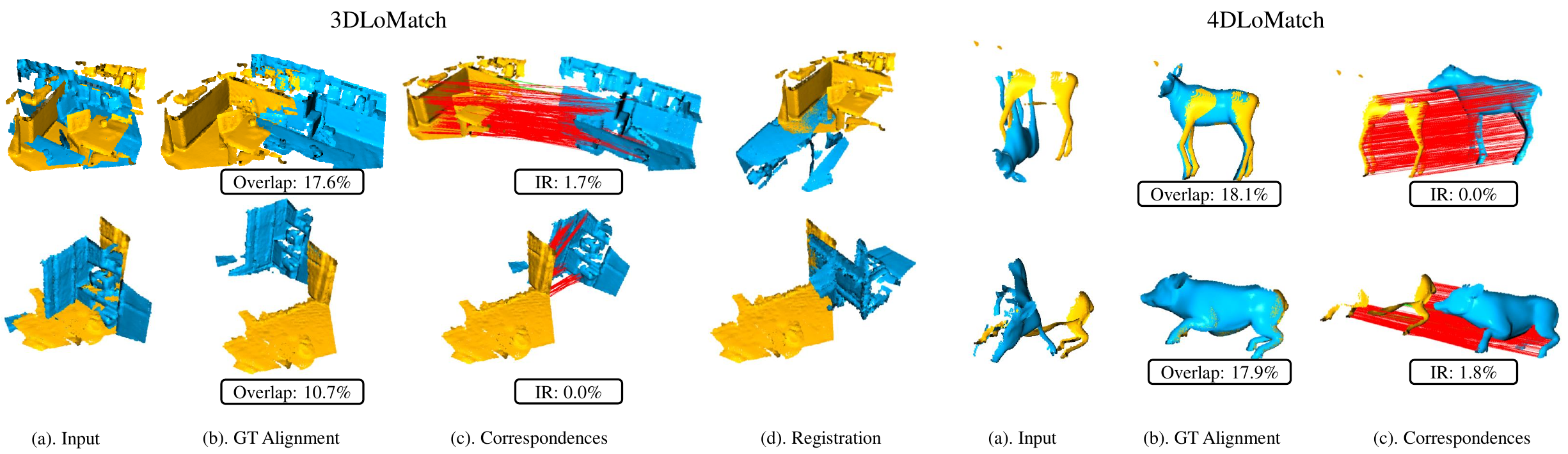}
  \caption{Failed cases on 3DLoMatch and 4DLoMatch.}
  \label{fig:failed_more}
\end{figure*}

\noindent\textbf{Further Discussion.}
Although \OURS{} achieves remarkable performance on both the rigid and non-rigid scenarios, we also notice the drawbacks of our method. The first is the efficiency of the attention mechanism. Although our local attention mechanism runs faster compared to that of Point Transformer~\cite{zhao2021point}, its running speed is still lower than that of convolutions, as shown in Tab.~\ref{tab:runtime}. Moreover, the intrinsic rotation invariance comes at the cost of losing the ability to match symmetric structures~(see the 4DLoMatch data of Fig.~\ref{fig:failed_more}). Furthermore, \OURS{} mainly relies on feature distinctiveness to implicitly filter out the occluded areas during the matching procedure, which makes it fail in cases with extremely limited overlap~(see the 3DLoMatch data of Fig.~\ref{fig:failed_more}). Finally, as normal data augmentation cannot work on intrinsically rotation-invariant methods, more data is required to train a larger model. 

\paragraph{Failure Cases.} We further show some failure cases in Fig.~\ref{fig:failed_more}. It can be observed that the failure on 3DLoMatch is caused by an extremely limited overlap on the flattened areas. In the first row, the overlap ratio is only $17.6\%$, and the overlap region is mainly on the floor. In the second row, the overlap region is even more limited~(with an overlap ratio of $10.7\%$) and mainly on a wall. For the 4DLoMatch, the failure is mainly due to the extremely limited overlap and the ambiguity caused by the symmetric structure. The first row shows a case with the two frames of point cloud showing a horse's left and right parts, with only $18.1\%$ overlap in the middle. The second row with $17.9\%$ overlap ratio also has a strong left-right ambiguity due to the symmetric structure of a pig, which accounts for many left-right mismatches.

{\small
\bibliographystyle{ieee_fullname}
\bibliography{egbib}

\begin{thebibliography}{10}\itemsep=-1pt

\bibitem{ao2021spinnet}
Sheng Ao, Qingyong Hu, Bo Yang, Andrew Markham, and Yulan Guo.
\newblock Spinnet: Learning a general surface descriptor for 3d point cloud
  registration.
\newblock In {\em CVPR}, 2021.

\bibitem{ba2016layer}
Jimmy~Lei Ba, Jamie~Ryan Kiros, and Geoffrey~E Hinton.
\newblock Layer normalization.
\newblock {\em arXiv preprint arXiv:1607.06450}, 2016.

\bibitem{bai2020d3feat}
Xuyang Bai, Zixin Luo, Lei Zhou, Hongbo Fu, Long Quan, and Chiew-Lan Tai.
\newblock D3feat: Joint learning of dense detection and description of 3d local
  features.
\newblock In {\em CVPR}, 2020.

\bibitem{barroso2020hdd}
Axel Barroso-Laguna, Yannick Verdie, Benjamin Busam, and Krystian Mikolajczyk.
\newblock Hdd-net: Hybrid detector descriptor with mutual interactive learning.
\newblock In {\em Proceedings of the Asian Conference on Computer Vision},
  2020.

\bibitem{birdal2015point}
Tolga Birdal and Slobodan Ilic.
\newblock Point pair features based object detection and pose estimation
  revisited.
\newblock In {\em 3DV}, 2015.

\bibitem{choy20194d}
Christopher Choy, JunYoung Gwak, and Silvio Savarese.
\newblock 4d spatio-temporal convnets: Minkowski convolutional neural networks.
\newblock In {\em CVPR}, 2019.

\bibitem{choy2019fully}
Christopher Choy, Jaesik Park, and Vladlen Koltun.
\newblock Fully convolutional geometric features.
\newblock In {\em ICCV}, 2019.

\bibitem{chua1997point}
Chin~Seng Chua and Ray Jarvis.
\newblock Point signatures: A new representation for 3d object recognition.
\newblock {\em IJCV}, 1997.

\bibitem{deng2018ppf}
Haowen Deng, Tolga Birdal, and Slobodan Ilic.
\newblock Ppf-foldnet: Unsupervised learning of rotation invariant 3d local
  descriptors.
\newblock In {\em ECCV}, 2018.

\bibitem{deng2018ppfnet}
Haowen Deng, Tolga Birdal, and Slobodan Ilic.
\newblock Ppfnet: Global context aware local features for robust 3d point
  matching.
\newblock In {\em CVPR}, 2018.

\bibitem{drost2010model}
Bertram Drost, Markus Ulrich, Nassir Navab, and Slobodan Ilic.
\newblock Model globally, match locally: Efficient and robust 3d object
  recognition.
\newblock In {\em CVPR}, 2010.

\bibitem{fischler1981random}
Martin~A Fischler and Robert~C Bolles.
\newblock Random sample consensus: a paradigm for model fitting with
  applications to image analysis and automated cartography.
\newblock {\em Communications of the ACM}, 1981.

\bibitem{gojcic2019perfect}
Zan Gojcic, Caifa Zhou, Jan~D Wegner, and Andreas Wieser.
\newblock The perfect match: 3d point cloud matching with smoothed densities.
\newblock In {\em CVPR}, 2019.

\bibitem{guo2013rotational}
Yulan Guo, Ferdous Sohel, Mohammed Bennamoun, Min Lu, and Jianwei Wan.
\newblock Rotational projection statistics for 3d local surface description and
  object recognition.
\newblock {\em IJCV}, 2013.

\bibitem{hou2021exploring}
Ji Hou, Benjamin Graham, Matthias Nie{\ss}ner, and Saining Xie.
\newblock Exploring data-efficient 3d scene understanding with contrastive
  scene contexts.
\newblock In {\em CVPR}, 2021.

\bibitem{hou2021pri3d}
Ji Hou, Saining Xie, Benjamin Graham, Angela Dai, and Matthias Nie{\ss}ner.
\newblock Pri3d: Can 3d priors help 2d representation learning?
\newblock In {\em ICCV}, 2021.

\bibitem{huang2016volumetric}
Chun-Hao Huang, Benjamin Allain, Jean-S{\'e}bastien Franco, Nassir Navab,
  Slobodan Ilic, and Edmond Boyer.
\newblock Volumetric 3d tracking by detection.
\newblock In {\em CVPR}, 2016.

\bibitem{huang2017tracking}
Chun-Hao~Paul Huang, Benjamin Allain, Edmond Boyer, Jean-S{\'e}bastien Franco,
  Federico Tombari, Nassir Navab, and Slobodan Ilic.
\newblock Tracking-by-detection of 3d human shapes: from surfaces to volumes.
\newblock {\em TPAMI}, 2017.

\bibitem{huang2021predator}
Shengyu Huang, Zan Gojcic, Mikhail Usvyatsov, Andreas Wieser, and Konrad
  Schindler.
\newblock Predator: Registration of 3d point clouds with low overlap.
\newblock In {\em CVPR}, 2021.

\bibitem{kingma2014adam}
Diederik~P Kingma and Jimmy Ba.
\newblock Adam: A method for stochastic optimization.
\newblock {\em arXiv preprint arXiv:1412.6980}, 2014.

\bibitem{li2022lepard}
Yang Li and Tatsuya Harada.
\newblock Lepard: Learning partial point cloud matching in rigid and deformable
  scenes.
\newblock In {\em CVPR}, 2022.

\bibitem{li20214dcomplete}
Yang Li, Hikari Takehara, Takafumi Taketomi, Bo Zheng, and Matthias
  Nie{\ss}ner.
\newblock 4dcomplete: Non-rigid motion estimation beyond the observable
  surface.
\newblock In {\em ICCV}, 2021.

\bibitem{newcombe2015dynamicfusion}
Richard~A Newcombe, Dieter Fox, and Steven~M Seitz.
\newblock Dynamicfusion: Reconstruction and tracking of non-rigid scenes in
  real-time.
\newblock In {\em CVPR}, 2015.

\bibitem{newcombe2011kinectfusion}
Richard~A Newcombe, Shahram Izadi, Otmar Hilliges, David Molyneaux, David Kim,
  Andrew~J Davison, Pushmeet Kohi, Jamie Shotton, Steve Hodges, and Andrew
  Fitzgibbon.
\newblock Kinectfusion: Real-time dense surface mapping and tracking.
\newblock In {\em IEEE International Symposium on Mixed and Augmented Reality},
  2011.

\bibitem{paszke2019pytorch}
Adam Paszke, Sam Gross, Francisco Massa, Adam Lerer, James Bradbury, Gregory
  Chanan, Trevor Killeen, Zeming Lin, Natalia Gimelshein, Luca Antiga, et~al.
\newblock Pytorch: An imperative style, high-performance deep learning library.
\newblock {\em NeurIPS}, 2019.

\bibitem{puy2020flot}
Gilles Puy, Alexandre Boulch, and Renaud Marlet.
\newblock Flot: Scene flow on point clouds guided by optimal transport.
\newblock In {\em ECCV}, 2020.

\bibitem{qi2017pointnet}
Charles~R Qi, Hao Su, Kaichun Mo, and Leonidas~J Guibas.
\newblock Pointnet: Deep learning on point sets for 3d classification and
  segmentation.
\newblock In {\em CVPR}, 2017.

\bibitem{qi2017pointnet++}
Charles~Ruizhongtai Qi, Li Yi, Hao Su, and Leonidas~J Guibas.
\newblock Pointnet++: Deep hierarchical feature learning on point sets in a
  metric space.
\newblock {\em NeurIPS}, 2017.

\bibitem{qin2022geometric}
Zheng Qin, Hao Yu, Changjian Wang, Yulan Guo, Yuxing Peng, and Kai Xu.
\newblock Geometric transformer for fast and robust point cloud registration.
\newblock In {\em CVPR}, 2022.

\bibitem{qin2023deep}
Zheng Qin, Hao Yu, Changjian Wang, Yuxing Peng, and Kai Xu.
\newblock Deep graph-based spatial consistency for robust non-rigid point cloud
  registration.
\newblock {\em arXiv preprint arXiv:2303.09950}, 2023.

\bibitem{rocco2018neighbourhood}
Ignacio Rocco, Mircea Cimpoi, Relja Arandjelovi{\'c}, Akihiko Torii, Tomas
  Pajdla, and Josef Sivic.
\newblock Neighbourhood consensus networks.
\newblock {\em NeurIPS}, 2018.

\bibitem{rusu2009fast}
Radu~Bogdan Rusu, Nico Blodow, and Michael Beetz.
\newblock Fast point feature histograms (fpfh) for 3d registration.
\newblock In {\em ICRA}, 2009.

\bibitem{rusu2008aligning}
Radu~Bogdan Rusu, Nico Blodow, Zoltan~Csaba Marton, and Michael Beetz.
\newblock Aligning point cloud views using persistent feature histograms.
\newblock In {\em IROS}, 2008.

\bibitem{saleh2020graphite}
Mahdi Saleh, Shervin Dehghani, Benjamin Busam, Nassir Navab, and Federico
  Tombari.
\newblock Graphite: Graph-induced feature extraction for point cloud
  registration.
\newblock In {\em 3DV}, 2020.

\bibitem{saleh2022cloudattention}
Mahdi Saleh, Yige Wang, Nassir Navab, Benjamin Busam, and Federico Tombari.
\newblock Cloudattention: Efficient multi-scale attention scheme for 3d point
  cloud learning.
\newblock {\em arXiv preprint arXiv:2208.00524}, 2022.

\bibitem{saleh2022bending}
Mahdi Saleh, Shun-Cheng Wu, Luca Cosmo, Nassir Navab, Benjamin Busam, and
  Federico Tombari.
\newblock Bending graphs: Hierarchical shape matching using gated optimal
  transport.
\newblock In {\em CVPR}, 2022.

\bibitem{sarlin2020superglue}
Paul-Edouard Sarlin, Daniel DeTone, Tomasz Malisiewicz, and Andrew Rabinovich.
\newblock Superglue: Learning feature matching with graph neural networks.
\newblock In {\em CVPR}, 2020.

\bibitem{sinkhorn1967concerning}
Richard Sinkhorn and Paul Knopp.
\newblock Concerning nonnegative matrices and doubly stochastic matrices.
\newblock {\em Pacific Journal of Mathematics}, 1967.

\bibitem{sun2021loftr}
Jiaming Sun, Zehong Shen, Yuang Wang, Hujun Bao, and Xiaowei Zhou.
\newblock Loftr: Detector-free local feature matching with transformers.
\newblock In {\em CVPR}, 2021.

\bibitem{sun2020circle}
Yifan Sun, Changmao Cheng, Yuhan Zhang, Chi Zhang, Liang Zheng, Zhongdao Wang,
  and Yichen Wei.
\newblock Circle loss: A unified perspective of pair similarity optimization.
\newblock In {\em CVPR}, 2020.

\bibitem{tang2022neural}
Jiapeng Tang, Lev Markhasin, Bi Wang, Justus Thies, and Matthias Nie{\ss}ner.
\newblock Neural shape deformation priors.
\newblock {\em arXiv preprint arXiv:2210.05616}, 2022.

\bibitem{tang2021learning}
Jiapeng Tang, Dan Xu, Kui Jia, and Lei Zhang.
\newblock Learning parallel dense correspondence from spatio-temporal
  descriptors for efficient and robust 4d reconstruction.
\newblock In {\em CVPR}, 2021.

\bibitem{thomas2019kpconv}
Hugues Thomas, Charles~R Qi, Jean-Emmanuel Deschaud, Beatriz Marcotegui,
  Fran{\c{c}}ois Goulette, and Leonidas~J Guibas.
\newblock Kpconv: Flexible and deformable convolution for point clouds.
\newblock In {\em ICCV}, 2019.

\bibitem{tombari2010unique}
Federico Tombari, Samuele Salti, and Luigi~Di Stefano.
\newblock Unique signatures of histograms for local surface description.
\newblock In {\em ECCV}, 2010.

\bibitem{trappolini2021shape}
Giovanni Trappolini, Luca Cosmo, Luca Moschella, Riccardo Marin, Simone Melzi,
  and Emanuele Rodol{\`a}.
\newblock Shape registration in the time of transformers.
\newblock {\em NeurIPS}, 2021.

\bibitem{vaswani2017attention}
Ashish Vaswani, Noam Shazeer, Niki Parmar, Jakob Uszkoreit, Llion Jones,
  Aidan~N Gomez, {\L}ukasz Kaiser, and Illia Polosukhin.
\newblock Attention is all you need.
\newblock {\em NeurIPS}, 2017.

\bibitem{wang2022you}
Haiping Wang, Yuan Liu, Zhen Dong, and Wenping Wang.
\newblock You only hypothesize once: Point cloud registration with
  rotation-equivariant descriptors.
\newblock In {\em ACM MM}, 2022.

\bibitem{wu2019pointpwc}
Wenxuan Wu, Zhiyuan Wang, Zhuwen Li, Wei Liu, and Li Fuxin.
\newblock Pointpwc-net: A coarse-to-fine network for supervised and
  self-supervised scene flow estimation on 3d point clouds.
\newblock {\em arXiv preprint arXiv:1911.12408}, 2019.

\bibitem{xie2020pointcontrast}
Saining Xie, Jiatao Gu, Demi Guo, Charles~R Qi, Leonidas Guibas, and Or Litany.
\newblock Pointcontrast: Unsupervised pre-training for 3d point cloud
  understanding.
\newblock In {\em ECCV}, 2020.

\bibitem{yang2018foldingnet}
Yaoqing Yang, Chen Feng, Yiru Shen, and Dong Tian.
\newblock Foldingnet: Point cloud auto-encoder via deep grid deformation.
\newblock In {\em CVPR}, 2018.

\bibitem{yew2022regtr}
Zi~Jian Yew and Gim~Hee Lee.
\newblock Regtr: End-to-end point cloud correspondences with transformers.
\newblock In {\em CVPR}, 2022.

\bibitem{yu2022riga}
Hao Yu, Ji Hou, Zheng Qin, Mahdi Saleh, Ivan Shugurov, Kai Wang, Benjamin
  Busam, and Slobodan Ilic.
\newblock Riga: Rotation-invariant and globally-aware descriptors for point
  cloud registration.
\newblock {\em arXiv preprint arXiv:2209.13252}, 2022.

\bibitem{yu2021cofinet}
Hao Yu, Fu Li, Mahdi Saleh, Benjamin Busam, and Slobodan Ilic.
\newblock Cofinet: Reliable coarse-to-fine correspondences for robust
  pointcloud registration.
\newblock {\em NeurIPS}, 2021.

\bibitem{yuan2020deepgmr}
Wentao Yuan, Benjamin Eckart, Kihwan Kim, Varun Jampani, Dieter Fox, and Jan
  Kautz.
\newblock Deepgmr: Learning latent gaussian mixture models for registration.
\newblock In {\em ECCV}, 2020.

\bibitem{zeng20173dmatch}
Andy Zeng, Shuran Song, Matthias Nie{\ss}ner, Matthew Fisher, Jianxiong Xiao,
  and Thomas Funkhouser.
\newblock 3dmatch: Learning local geometric descriptors from rgb-d
  reconstructions.
\newblock In {\em CVPR}, 2017.

\bibitem{zhang2022pcr}
Yu Zhang, Junle Yu, Xiaolin Huang, Wenhui Zhou, and Ji Hou.
\newblock Pcr-cg: Point cloud registration via deep explicit color and
  geometry.
\newblock In {\em ECCV}, 2022.

\bibitem{zhao2021point}
Hengshuang Zhao, Li Jiang, Jiaya Jia, Philip~HS Torr, and Vladlen Koltun.
\newblock Point transformer.
\newblock In {\em ICCV}, 2021.

\end{thebibliography}
}

\end{document}